\documentclass[lettersize,journal]{IEEEtran}

\usepackage{amsmath,amsfonts}
\usepackage{algorithmic}
\usepackage{array}
\usepackage[caption=false,font=normalsize,labelfont=sf,textfont=sf]{subfig}
\usepackage{textcomp}
\usepackage{stfloats}
\usepackage{url}
\usepackage{verbatim}
\usepackage{graphicx}
\usepackage{balance}
\usepackage{cite}

\usepackage[table]{xcolor}
\usepackage{orcidlink}
\usepackage[utf8]{inputenc}
\usepackage[english]{babel}
\usepackage{hyperref}
\usepackage{acronym}
\usepackage{soul}
\usepackage{caption}
\usepackage{multirow}

\acrodef{AI}[AI]{Artificial Intelligence}
\acrodef{FPGA}[FPGA]{Field-Programmable Gate Array}
\acrodef{N-MNIST}[N-MNIST]{Neuromorphic-MNIST}
\acrodef{SHD}[SHD]{Spiking Heidelberg Digits}
\acrodef{BP}[BP]{Backpropagation}
\acrodef{ANN}[ANN]{Artificial Neural Network}
\acrodefplural{ANN}[ANNs]{Artificial Neural Networks}
\acrodef{RNN}[RNN]{Recurrent Neural Network}
\acrodef{RSNN}[RSNN]{Recurrent Spiking Neural Network}
\acrodefplural{RNN}[RNNs]{Recurrent Neural Networks}
\acrodef{BPTT}[BPTT]{Back-Propagation Through Time}
\acrodef{ETLP}[ETLP]{Event-based Three-factor Local Plasticity}
\acrodef{DRTP}[DRTP]{Direct Random Target Projection}
\acrodef{SNN}[SNN]{Spiking Neural Network}
\acrodefplural{SNN}[SNNs]{Spiking Neural Networks}
\acrodef{LIF}[LIF]{Leaky-Integrated and Fire}
\acrodef{ALIF}[ALIF]{Adaptive Leaky-Integrate and Fire}
\acrodef{MSE}[MSE]{Mean Squared Error}
\acrodef{RTRL}[RTRL]{Real-Time Recurrent Learning}
\acrodef{ASIC}[ASIC]{Application-Specific Integrated Circuit}
\acrodef{MAML}[MAML]{Model-Agnostic Meta-Learning}

\date{\today}

\begin{document}

\title{ETLP: Event-based Three-factor Local Plasticity for online learning with neuromorphic hardware}

\author{
Fernando M. Quintana \orcidlink{0000-0001-5042-9399}, 
Fernando Perez-Peña \orcidlink{0000-0003-3586-2930} \IEEEmembership{Member, IEEE}, 
Pedro L. Galindo \orcidlink{0000-0003-0892-8113}, \\
Emre O. Neftci \orcidlink{0000-0002-0332-3273} \IEEEmembership{Member, IEEE}, 
Elisabetta Chicca \orcidlink{0000-0002-5518-8990} \IEEEmembership{Senior Member, IEEE}, 
Lyes Khacef \orcidlink{0000-0002-4009-174X}

\thanks{Fernando M. Quintana, Fernando Perez-Peña, and Pedro L. Galindo are with School of Engineering, University of Cádiz, Avda. Universidad de Cádiz, nº 10, Puerto Real, 11519, Cádiz, Spain (e-mail: fernando.quintana@uca.es, pedro.galindo@uca.es, fernandoperez.pena@uca.es).}
\thanks{Emre O. Neftci is with the Neuromorphic Software Ecosystems, Peter Grünberg Institute, Forschungszentrum Jülich, Technologie Zentrum Aachen Dennewartstraße 25 – 27, Aachen, 52068, Germany (e-mail: e.neftci@fz-juelich.de).}
\thanks{Elisabetta Chicca and Lyes Khacef are with Bio-inspired Circuits and Systems (BICS), University of Groningen, Nijenborgh 4, Groningen, NL-9747AG, the Netherlands (e-mail: e.chicca@rug.nl, l.khacef@rug.nl).}
}

%\markboth{IEEE TRANSACTIONS ON NEURAL NETWORKS AND LEARNING SYSTEMS, Vol. 1, No. 1, January 2021}
%{Fernando M. Quintana \MakeLowercase{\textit{(et al.)}}: ETLP: Event-based Three-factor Local Plasticity for online learning with neuromorphic hardware.}

\maketitle

% ------------------------------------------------------

\begin{abstract}
Neuromorphic perception with event-based sensors, asynchronous hardware and spiking neurons is showing promising results for real-time and energy-efficient inference in embedded systems.
The next promise of brain-inspired computing is to enable adaptation to changes at the edge with online learning. 
However, the parallel and distributed architectures of neuromorphic hardware based on co-localized compute and memory imposes locality constraints to the on-chip learning rules.
We propose in this work the Event-based Three-factor Local Plasticity (ETLP) rule that uses (1) the pre-synaptic spike trace, (2) the post-synaptic membrane voltage and (3) a third factor in the form of projected labels with no error calculation, that also serve as update triggers.
We apply ETLP with feedforward and recurrent spiking neural networks on visual and auditory event-based pattern recognition, and compare it to Back-Propagation Through Time (BPTT) and eProp.
We show a competitive performance in accuracy with a clear advantage in the computational complexity for ETLP.
We also show that when using local plasticity, threshold adaptation in spiking neurons and a recurrent topology are necessary to learn spatio-temporal patterns with a rich temporal structure.
Finally, we provide a proof of concept hardware implementation of ETLP on FPGA to highlight the simplicity of its computational primitives and how they can be mapped into neuromorphic hardware for online learning with low-energy consumption and real-time interaction.
\end{abstract}


% ------------------------------------------------------

\section{Introduction}
\label{sec:introduction}
\IEEEPARstart{T}{he} quest for \textit{intelligence} in embedded systems is becoming necessary to sustain the current and future developments of edge devices. 
Specifically, the rapid growth of wearable and connected devices brings new challenges for the analysis and classification of the streamed data at the edge, while emerging autonomous systems require adaptability in many application areas such as brain–machine interfaces, AI-powered assistants or autonomous vehicles. 
In addition, the current trend points toward increased autonomy and closed-loop feedback, where the results of the data analysis are translated into immediate action within a very tight latency window \cite{Rabaey_etal19}. 
Hence, in addition to input (sensors) and output (actuators) modalities, those systems need to combine understanding, reasoning and decision-making \cite{Rabaey_etal19}, and off-line training techniques cannot account for all possible scenarios. 
Consequently, there is a need for online learning in embedded systems under severe constraints of low latency and low power consumption.

At the same time, \ac{AI} based on deep learning has led to immense breakthroughs in many applications areas such as object detection/recognition and natural language processing thanks to the availability of labeled databases and the increasing computing power of standard CPU and GPU hardware. 
However, the sources of this increasing computing performance have been challenged by the end of Dennard scaling \cite{Dennard_etal74,Dennard_etal07} in around 2005, resulting in the inability to increase clock frequencies significantly. Today, advances in silicon lithography still enable the miniaturization of electronics, but as transistors reach atomic scale and fabrication costs continue to rise, the classical technological drive that has supported Moore’s Law for fifty years is reaching a physical limit and is predicted to flatten by 2025 \cite{Shalf20}.
Hence, deep learning progress with current models and implementations will be constrained by its computational requirements, and will be technically, economically and environmentally unsustainable \cite{Thompson_etal20,Thompson_etal21}.
The way forward will thus require simultaneous paradigm shifts in the computational concepts and models as well as the hardware architectures, by specializing the architecture to the task \cite{Shalf20,Christensen_etal22}.

Therefore, in contrast to standard von Neumann computing architectures where processing and memory are separated, neuromorphic architectures get inspiration from the biological nervous system and propose parallel and distributed implementations of synapses and neurons where computing and memory are co-localized \cite{Mead_Conway80,Chicca_etal14b}. 
Neuromorphic implementations have grown in the past thirty years to cover a wide range of hardware \cite{Schuman_etal17,Basu_etal22}, from novel materials and devices \cite{Boybat_etal18,Payvand_etal22} to analog \cite{Indiveri_etal11,Qiao_etal15,Chicca_etal14b} and digital CMOS technologies \cite{Frenkel_etal19,Stuijt_etal21,Muliokov_etal21}.
Recent works have shown important gains in computation delay and energy-efficiency when combining event-based sensing (sensor level), asynchronous processing (hardware level) and spike-based computing (algorithmic level) in neuromorphic systems \cite{ceolini_etal20,Davies_etal21,Muller-Cleve_etal22}. 
On the other hand, adaptation with online learning is still ongoing research because neuromorphic hardware constraints prevent the use of exact gradient-based optimization with \ac{BP} and impose the use of local plasticity \cite{Neftci_etal19,Zenke_Neftci21,Khacef_etal22}.

Gradient-based learning \cite{LeCun_etal98} has been recently applied to offline training of \acp{SNN}, where the \ac{BPTT} algorithm coupled with surrogate gradients is used to solve the (1) temporal credit assignment by unrolling the \ac{SNN} like in standard \acp{RNN} \cite{Neftci_etal19}, as well the (2) spatial credit assignment by assigning the ``blame'' to each neuron across the layers with respect to the objective function to optimize. These operations require global computations that are not biologically plausible \cite{Bengio_etal15} and do not fit with online learning in neuromorphic hardware. Specifically:

\begin{itemize}
    \item \ac{BPTT} is non-local in time because it keeps all the network activities for the duration of each sample, which requires a large memory footprint, and it operates in two phases, which adds latency in the backward pass.
    \item \ac{BPTT} is non-local in space because it back-propagates the gradient back from the output layer to all the other layers. It adds further latency and creates a locking effect \cite{Czarnecki_etal17}, since a synapse cannot be updated until the feedforward pass, error evaluation and backward pass are completed. It also creates a weight transport problem \cite{Lillicrap_etal16} since the error signal arriving at some neuron in a hidden layer must be multiplied by the strength of that neuron’s forward synaptic connections to the source of the error, which requires extra circuitry and consumes energy \cite{Kraiser_etal20}.
\end{itemize}

The lack of locality in time and space makes \ac{BPTT} unsuitable for the hardware constraints of neuromorphic implementations for which latency, efficiency and scalability are of primary concern.

Recently, intensive research in neuromorphic computing has been dedicated to bridge the gap between gradient-based learning \cite{LeCun_etal98,LeCun_etal15} and local synaptic plasticity \cite{McNaughton_etal78,Gerstner_etal93,Stuart_Sakmann94,Markram_etal95,Morrison_etal08} by reducing the non-local information requirements. However, these local plasticity rules often resulted in a cost in accuracy for complex pattern recognition tasks \cite{Eshraghian_etal21}.
On the one hand, temporal credit assignment can be approximated by using eligibility traces \cite{Zenke_Ganguli18,Bellec2020} that solve the distal reward problem by bridging the temporal gap between the network output and the feedback signal that may arrive later in time \cite{Izhikevich07}. 
On the other hand, several strategies are explored for solving the spatial credit assignment by using feedback alignment \cite{Lillicrap_etal16}, direct feedback alignment \cite{Nokland16} and random error \ac{BP} \cite{Neftci_etal17}. These approaches partially solve the spatial credit assignment problem \cite{Eshraghian_etal21}, since they still suffer from the locking effect. This constraint is further relaxed by replacing the global loss by a number of local loss functions \cite{Mostafa_etal18,Neftci_etal19,Kraiser_etal20} or using \ac{DRTP} \cite{Frenkel_etal21b}. The method of local losses solves the locking problem, but is also a less effective spatial credit assignment mechanism compared to \ac{BP}.

In this work, we present the \ac{ETLP} rule for for online learning scenarios with spike-based neuromorphic hardware where a teaching signal can be provided to the network. 
The objective is to enable intelligent embedded systems to adapt to new classes or new patterns and improve their performance after deployment. 
\ac{ETLP} is local in time by using spike traces \cite{Bellec2020}, and local in space by using \ac{DRTP} \cite{Frenkel_etal21b} which means that no error is explicitly calculated. Instead, the labels are directly provided to each neuron in the form of a third-factor that triggers the weight update in an asynchronous way.
We present in the next section the algorithmic formulation of \ac{ETLP}, then apply it to visual and auditory event-based pattern recognition problems with different levels of temporal information. We explore the impact of an adaptive threshold mechanism in spiking neurons as well as explicit recurrence in the network, and compare \ac{ETLP} to eProp \cite{Bellec2020} and \ac{BPTT} in terms of accuracy and computational complexity. Finally, we demonstrate a proof of concept hardware implementation of \ac{ETLP} on FPGA, and discuss possible improvements for online learning in real-world applications.

% ------------------------------------------------------

\section{Methods}
\noindent In order to achieve online learning based solely on local information, \ac{ETLP} is divided into two main parts: (1) the online calculation of the gradient to solve the temporal dependency as previously done in other works such as eProp \cite{Bellec2020} and DECOLLE \cite{Kraiser_etal20}, and (2) the update of the gradient with \ac{DRTP} \cite{Frenkel2021} to solve the spatial dependency by directly using the labels instead of a global error calculation.

\subsection{Neuron and synapse model}
The spiking neuron model used in this paper is the \ac{ALIF} with a soft-reset mechanism proposed in \cite{Bellec2020}.

\begin{subequations}\label{eq:LIF}
\begin{align}
    A_j(t) &= v_{th} + \theta a_j(t) \label{eq:LIFa}\\
    a_j(t) &= \gamma a_j(t-1) + s_j(t-1) \label{eq:LIFb}\\
    v_j(t) &= \alpha v_j(t-1) + \sum_i W_{ij}x_i(t)-s_j(t-1)v_{th} \label{eq:LIFc}\\
    s_j(t) &= H(v_j(t)-A_j(t)) \label{eq:LIFd}
\end{align}
\end{subequations}

Where $v_j$ describes the voltage of neuron $j$ with a decay constant $\alpha$, $W_{ij}$ the synaptic weight between neuron $j$ and input $i$, $x_i$ the input from the neuron $i$ of the previous layer, $s_j$ the spike state (0 or 1) of the neuron $j$, $H$ the activation function, that in our case is the step function and $A_j$ the instantaneous threshold obtained from the baseline threshold ($v_{th}$) and a trace ($a_j$) produced by the activity of neuron $j$, with a scaling factor $\theta$ and a decay constant $\gamma$ .

\subsection{Temporal dependency}
\label{sec:gradient}
\begin{figure}[!t]
  \centering
  \includegraphics[width=3.5in]{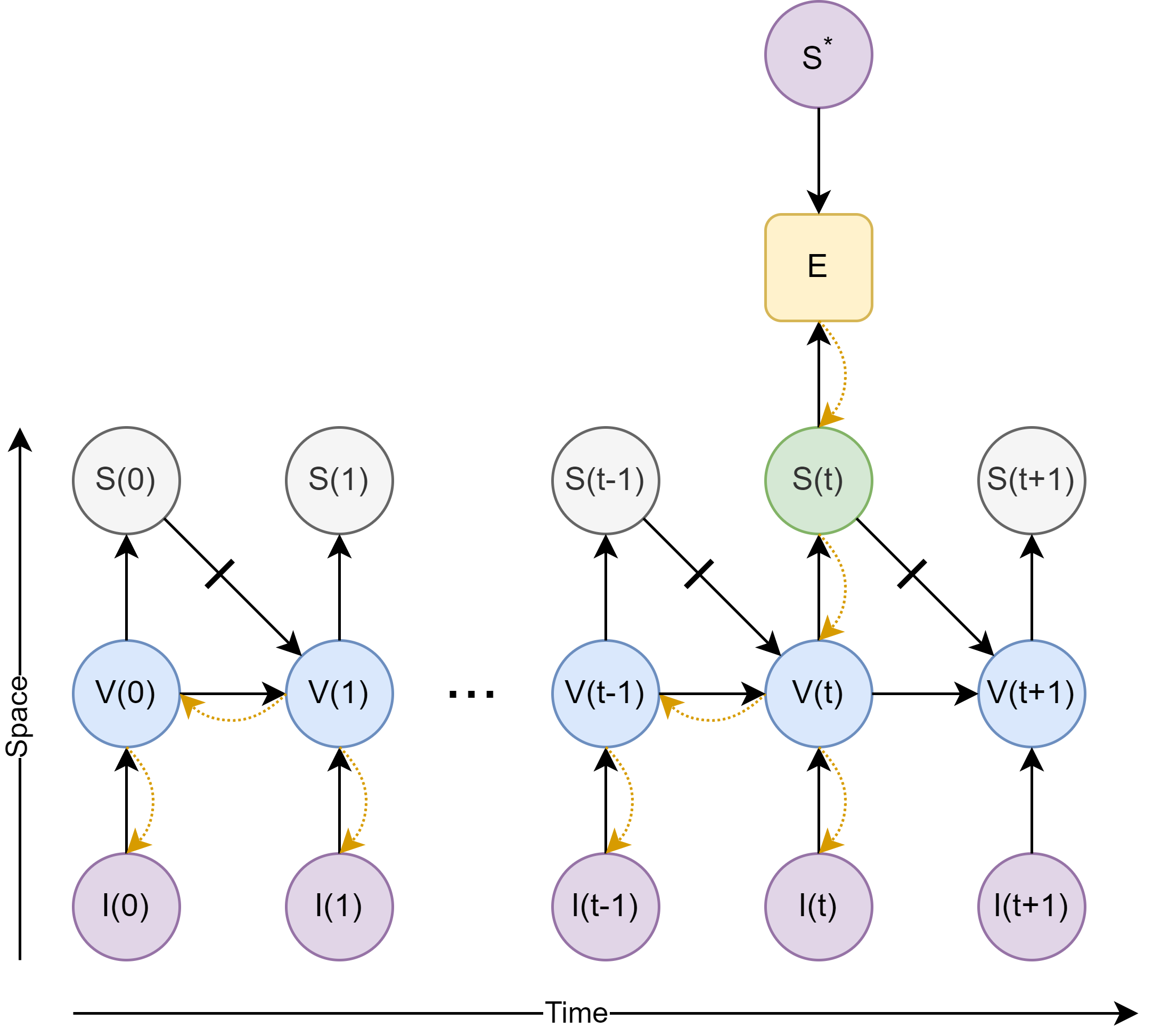}
  \caption{Computational graph for a 5 time steps simulation on a 1-layer \ac{SNN} using soft-reset \ac{LIF} neurons and back-propagation through time. Where I, V and S are vectors representing the neuron input, voltage and output spike, respectively. $\text{S}^*$ is the target output and E is the error. The black arrows represent the forward pass of the data, while the curved orange arrows represent the gradient backward pass. The neuron resets is detached from the computational graph in the gradient calculation.}
  \label{fig:computational_graph}
\end{figure}

Typically, the training of \ac{SNN} is based on many-to-many (sequence-to-sequence) models, where an error is calculated at each time step and then the gradient is calculated based on the combination of these errors \cite{Zenke_Neftci21}. In our case, in order to perform an event-driven update, we have used a many-to-one model where only the result of the network obtained at a specific time step is taken into account. This is shown in the computational graph in Fig. \ref{fig:computational_graph}, where the gradient is computed at timestep $t$. The gradient obtained from the \ac{ALIF} has only the addition of a threshold-dependent trace in the pre-synaptic spike trace, compared to a \ac{LIF} \cite{Bellec2020}. Therefore, for simplicity in this section, the calculation of the gradient is based on a \ac{LIF} neuron.

\begin{figure*}[!t]
\begin{subequations}
\label{eq:gradient_comp}
\begin{align}
    \frac{dE(t)}{dW} &= \frac{dE(t)}{dV(t)}I(T) + \frac{dE(t)}{dV(t)}\frac{dV(t)}{dV(t-1)}I(t-1) + \dots + \frac{dE(t)}{dV(t)}\frac{\partial V(t)}{\partial V(t-1)}\dots\frac{\partial V(1)}{\partial V(0)}I(0)\label{eq:gradient_comp_a}\\
    \frac{dE(t)}{dW} &= (S(t) - S^*)\varphi(t)I(t) + \dots + (S(t) - S^*)\varphi(t)~I(0)\alpha^t\label{eq:gradient_comp_b}\\
    \frac{dE(t)}{dW} &= (S(t) - S^*)\varphi(t)\underbrace{( I(t) + \alpha I(t-1) + \alpha^2 I(t-2) + \dots + \alpha^tI(0))}_{\varepsilon_{pre}(t)=\sum_{i=0}^t \alpha^{t-i}I(i)}\label{eq:gradient_comp_c}\\
    \frac{dE(t)}{dW} &= \underbrace{\color{blue}{\varepsilon_{pre}(t)}\color{green}{\varphi(t)}}_{e(t) = \varepsilon_{pre}(t)\varphi(t)}\color{red}{(S(t) - S^*)}\label{eq:gradient_comp_d}
\end{align}
\end{subequations}
\end{figure*}

\begin{subequations}
\begin{align}
    \begin{split}
    \frac{dE}{dW} ={}&\frac{\partial E}{\partial V(t)}\frac{\partial V(t)}{\partial W} + \frac{\partial E}{\partial V(t-1)}\frac{\partial V(t-1)}{\partial W} + \ldots + \\ &+ \frac{\partial E}{\partial V(0)}\frac{\partial V(0)}{\partial W}
    \end{split}\\
    \begin{split}
    \frac{dE}{dW} ={}&\frac{\partial E}{\partial V(t)}I(t) + \frac{\partial E}{\partial V(t-1)}I(t-1) + \ldots +\\&+\frac{\partial E}{\partial V(0)}I(0)
    \end{split}
    \label{eq:dE_dW}
\end{align}
\end{subequations}

In classification problems, the cross-entropy loss function is often used. However, as this depends on the softmax activation function which is not local (it depends on all output neurons values for the final calculation), we have used the \ac{MSE}. Consequently, the gradient of the error with respect to the network weights (according to equation \ref{eq:LIF}) would be as on equation \ref{eq:dE_dW}

Following equation \ref{eq:LIF}, the gradients of the network are as follows:

\begin{equation}
\begin{alignedat}{2}
&\frac{\partial S(t)}{\partial V(t)} = \phi(V(t)-A(t)) = \varphi(t)~&~&\frac{\partial V(t)}{\partial V(t-1)} = \alpha\\
&\frac{\partial E}{\partial S(t)} = S(t) - S^*~&~&\frac{\partial V(t)}{\partial W} = I(t)
\end{alignedat}
\end{equation}

where $\phi(x)$ is the surrogate gradient of the Heaviside function (H), $S^*$ the target output, $S(t)$ the output spike of the neuron and I(t) the input at time $t$. 

Assuming that the error is calculated at time t=T and calling $\frac{dE(t)}{dV(t)}=(S(t) - S^*)$, the gradient of the error with respect to the weights in a period of time T is defined as an equation of three factors (equation \ref{eq:gradient_comp_d}) which we will call: (1) pre-synaptic spike trace (blue), (2) surrogate gradient of the post-synaptic voltage (green) and (3) external signal (red). The pre-synaptic spike trace can be expressed as a low-pass filter of the input spikes to the neuron at each synapse, derived from equation \ref{eq:gradient_comp_c} as formulated in equation \ref{eq:e_trace}.

\begin{equation}
\label{eq:e_trace}
    \varepsilon_{pre}(t) = \alpha \varepsilon_{pre}(t-1) + I(t)
\end{equation}

Since the threshold adaptation must be taken into account for the gradient computation of the \ac{ALIF} model, the product of (1) the pre-synaptic spike trace and (2) the surrogate gradient of the post-synaptic voltage is modified according to equation \ref{eq:e_trace_full} \cite{Bellec2020}, where $e(t)$ is a function of (1) the pre-synaptic spike trace $\varepsilon(t)$, (2) the surrogate gradient of the post-synaptic voltage $\psi$ and (3) the threshold adaptation trace $a(t)$ at time step $t$, $\theta$ is the increase in the adaptive threshold, and $\gamma$ is the threshold decay constant.

\begin{align}
\label{eq:e_trace_full}
    \varepsilon_{adapt}(t) &= \varepsilon_{pre}(t)\varphi(t) + (\gamma - \varphi(t)\theta)\varepsilon_{adapt}(t-1)\\
    e(t) &= \varphi(t) (\varepsilon_{pre}(t) - \theta \varepsilon_{adapt}(t))
\end{align}

\subsection{Spatial dependency}
The back-propagation of the errors in calculating the network gradients means that there is a spatial dependency in training. This dependency results in non-local update to each neuron or synapse for two main reasons: (1) weight update locking and (2) weight transport problem as explained in section \ref{sec:introduction}).

\ac{DRTP} solves this problem by using one-hot-encoded targets in supervised classification problems as a proxy for the sign of the error \cite{Frenkel2021}. Therefore, instead of waiting for the data to propagate forward through the layers to calculate the global error which would be then back-propagated, the targets (i.e. labels) are used directly. Together with feedback alignment using fixed random weights from the teaching neurons in the gradient computation, \ac{DRTP} solves both spatial dependency problems (i.e. update locking and weight transport) \cite{Frenkel2021}. 
Since there is no propagation of gradient information between layers, the gradient computation is performed in a similar way in each layer, so the computational graph (Fig. \ref{fig:computational_graph}) is the same across layers. The difference is that, instead of using the gradient of the error, the projected labels are used.
As a result, in contrast with eProp, there is no need for a synaptic eligibility trace which we define as a low-pass filter of the Hebbian component (i.e. pre- and pos-synaptic information) of the learning rule \cite{Gerstner_etal18}.

\begin{figure}[!t]
    \centering
    \includegraphics[width=3.5in]{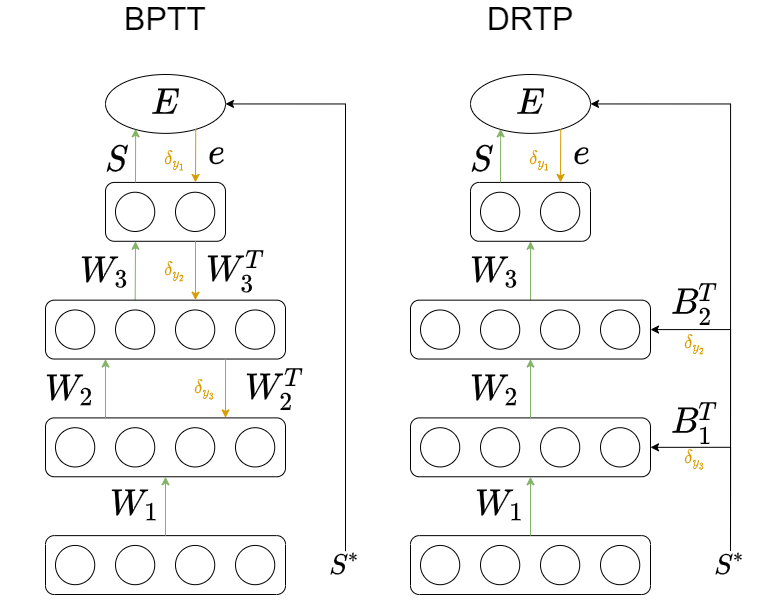}
    \caption{Weight update mechanism on \ac{BPTT} and \ac{DRTP}. In the case of \ac{BPTT}, and error is calculated using the output layer and the target $S^*$, the error is then propagated backward through the layers using the transpose of the synaptic weights. On \ac{DRTP}, the error is only calculated at the output layer, the weights in the hidden layer are updated using directly the target, using random fixed matrix.}
    \label{fig:DRTP}
\end{figure}

\subsection{ETLP}
\ac{ETLP} is based on a many-to-one learning architecture \cite{Zenke_Neftci21}, where the error is calculated at a specific time step of an input streaming data (Fig. \ref{fig:computational_graph}), that can be defined as a random point in an input sequence of spikes, as defined in section \ref{sec:gradient}. Learning is made using target labels with one-hot-encoding, representing the targets as teaching neurons.
Since the targets are represented as a one-hot-encoded vector, and the learning algorithm is updated using a many-to-one architecture (as exposed in section \ref{sec:gradient}), they can be represented as teaching neurons that fire an event at the end of a time window. The teaching neurons have synapses that connect them to the output neurons and the neurons in the hidden layers, firing at a certain frequency (e.g. $100\ Hz$), obtaining a fully event-based supervised learning technique.

\subsubsection{Synapse traces}
In the case of an \ac{ALIF} neuron, each synapse requires two traces: (1) the pre-synaptic spike trace (equation \ref{eq:pre_trace}) which holds the pre-synaptic activity information, where $I(t)$ represent the input spike and $\alpha$ the neuron time constant, and (2) the threshold adaptation trace (equation \ref{eq:adap_trace}) which holds the threshold value depending on the post-synaptic activity. 
In the case of a \ac{LIF}, only a trace of the pre-synaptic spikes are required. 
Finally, $e(t)$ in equation \ref{eq:e_trace_2} consists of a combination of the pre-synaptic spike trace, the surrogate gradient of the post-synaptic voltage (equation \ref{eq:post_trace}) and the threshold adaptation trace when using the \ac{ALIF} neuron.

\begin{align}
    \varepsilon_{pre}(t) &= \alpha \varepsilon_{pre}(t-1) + I(t) \label{eq:pre_trace}\\
    \varphi(t) &= \gamma max\left(0, 1 - \left|V(t)-A(t)\right|\right) \label{eq:post_trace}\\
    \varepsilon_{adapt}(t) &= \varepsilon_{pre}(t)\varphi(t) + (\gamma - \varphi(t)\theta)\varepsilon_{adapt}(t-1) \label{eq:adap_trace}\\
    e(t) &= \varphi(t) (\varepsilon_{pre}(t) - \theta \varepsilon_{adapt}(t)) \label{eq:e_trace_2}
\end{align}

The update of synapses is event-driven, i.e., each time the postsynaptic (hidden or output) neuron receives an input spike from a master neuron. In the case of hidden layers, the update is performed based on the equation \ref{eq:weight_update_hid}, where $I$ represents the input from the teaching neurons and $e$ the product of surrogate gradient and pre-synaptic and threshold adaptation traces. Thus, by using only local information per synapse (pre-synaptic and post-synaptic values), as well as a firing signal from a neuron, learning only takes place when an activation signal is received from the learning neurons.

\subsubsection{Synaptic plasticity}
The synapse update is event-driven, i.e., each time the post-synaptic neuron (hidden or output) receives a spike from a teaching neuron. In the case of the hidden layers, the update is made based on equation \ref{eq:weight_update_hid}, where $I$ represent the input from the teaching neurons. Thus, \ac{ETLP} learning uses local information per synapse (pre-synaptic spike trace, surrogate gradient of the post-synaptic voltage and teaching neurons weights) and is triggered when a spike from a teaching neuron is received. Otherwise, the weights of the network are not updated.
For the output layer, two different types of connections are made between the teaching neurons and the output layer. Each teaching neuron is connected in a one-to-one fashion with an excitatory synapse with the output neurons and with the other output neurons with an inhibitory synapse. As a result, the output neuron receive an input current $I$ that each time it receive an input spike from the teaching neuron, applying equation \ref{eq:weight_update_out}, where $e_{out}$ is the function over the pre-synaptic spike trace, the surrogate gradient of the post-synaptic voltage and the threshold adaptation trace of the output neuron, and $s_{out}$ is the spike of the output neuron.

\begin{equation}
\label{eq:weight_update_hid}
    W(t+1)_{in/rec} = W(t)_{in/rec} + \gamma I(t)e(t)_{in/rec}\\
\end{equation}

\begin{equation}
\label{eq:weight_update_out}
    W(t+1)_{out} = W(t)_{out} + \gamma \frac{2s(t)_{out} - I(t) - 1}{2} e(t)_{out}
\end{equation}

\begin{figure}[!t]
    \centering
    \includegraphics[width=3.5in]{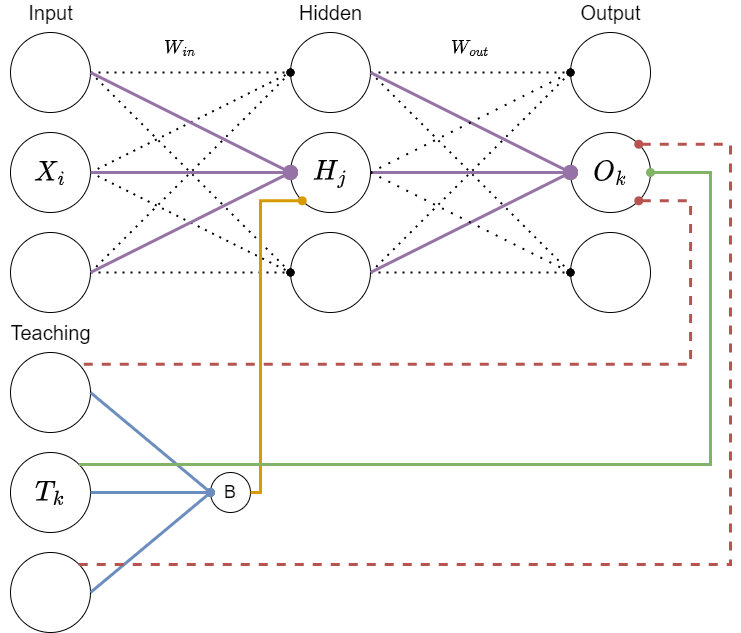}
    \caption{Feedforward \ac{SNN} with \ac{ETLP} learning. The network consists of a hidden layer, an output layer and teaching neurons. The output layer is connected to the teaching neurons by both excitatory (green) and inhibitory (dashed red) synapses. The learning layer is also connected to the group of recurrent neurons via a random weight matrix B. When a class is active, the neuron in the learning layer starts to emit spikes at a certain frequency (e.g. 100Hz), thus modifying the synapses of the neurons according to the equations \ref{eq:weight_update_hid} and \ref{eq:weight_update_out}.}
\end{figure}

% ------------------------------------------------------

\section{Results}
\noindent To test the performance of \ac{ETLP}, the rule was compared with \ac{BPTT} (non-local in time and space) and eProp (non-local in space) on two different datasets.
The final accuracy calculation is based on a rate decoding (i.e. winning neuron is the one firing most during the sample presentation) of the final layer's output spikes.

The first dataset is \ac{N-MNIST} \cite{orchard2015converting}, recorded by moving the ATIS sensor over MNIST samples (10 classes) in three saccades of 100ms each. The resulting information is therefore mainly spatial, where the temporal structure is exclusively related to the movements of the sensor. 
The second dataset is \ac{SHD} \cite{cramer2020heidelberg}, a set of 10000 spoken digit audios from 0 to 9 in both English and German (20 classes). The dataset was generated using an artificial cochlea, producing spikes in 700 input channels. In contrast with \ac{N-MNIST}, \ac{SHD} contains both a spatial and a rich temporal structure \cite{Perez-Nieves_etal21}.

For \ac{N-MNIST}, we used only the first saccade of about 100ms with a time step of  of 1ms, resulting in 100 time steps per sample.
For \ac{SHD}, we limited the samples duration to 1s with a time step of 10ms to result in 100 time steps per sample, similar to \ac{N-MNIST}. This time binning provides enough temporal information to reach a state-of-the-art performance in accuracy \cite{Bouanane_etal22}.

The hyper-parameters used on the simulations for both datasets are shown in Tab. \ref{tab:hyperparameters}, where dt is the algorithmic timestep, lr the learning rate, $\tau_m$ the membrane decay time constant, $\tau_a$ the threshold adaptation decay time constant, $v_{th}$ the base threshold. 
The PyTorch software implementation is publicly available on Github \footnote{\href{https://github.com/ferqui/ETLP}{https://github.com/ferqui/ETLP}}.
All the results presented in this section are averaged over three runs with different random initialization of the synaptic weights.

\begin{table}
    \centering
    \caption{Hyperparameter used on \ac{SHD} and \ac{N-MNIST} datasets.}
    \label{tab:hyperparameters}
    \begin{tabular}{|c|c|c|}
        \hline
        \textbf{Parameter} & \textbf{\ac{N-MNIST}} & \textbf{\ac{SHD}} \\
        \hline
        dt & 1ms & 10ms \\
        \hline
        lr & $5e^{-4}$ & $5e^{-4}$ \\
        \hline
        $\tau_m$ & 80ms & 1000ms \\
        \hline
        $\tau_a$ & 10ms & 1000ms \\
        \hline
        $v_{th}$ & 1 & 1 \\
        \hline
        batch size & 128 & 128 \\
        \hline
        Hidden size & 200 & 450 \\
        \hline
        Refactory period (time steps)& 5 & 5 \\
        \hline
    \end{tabular}
\end{table}

% ----------------- N-MNIST ----------------- %

\subsection{N-MNIST}
For the \ac{N-MNIST} dataset, a feedforward network of \ac{LIF} neurons with 2064 input neurons ($32 \times 32 = 1032$ neurons for each polarity), 200 hidden neurons and 10 output neurons was used. 
Indeed, explicit recurrence is not needed when the information in the data are mostly spatial \cite{Bouanane_etal22}.

\begin{figure*}[!h]
\centering
\subfloat[]{\includegraphics[width=0.49\textwidth]{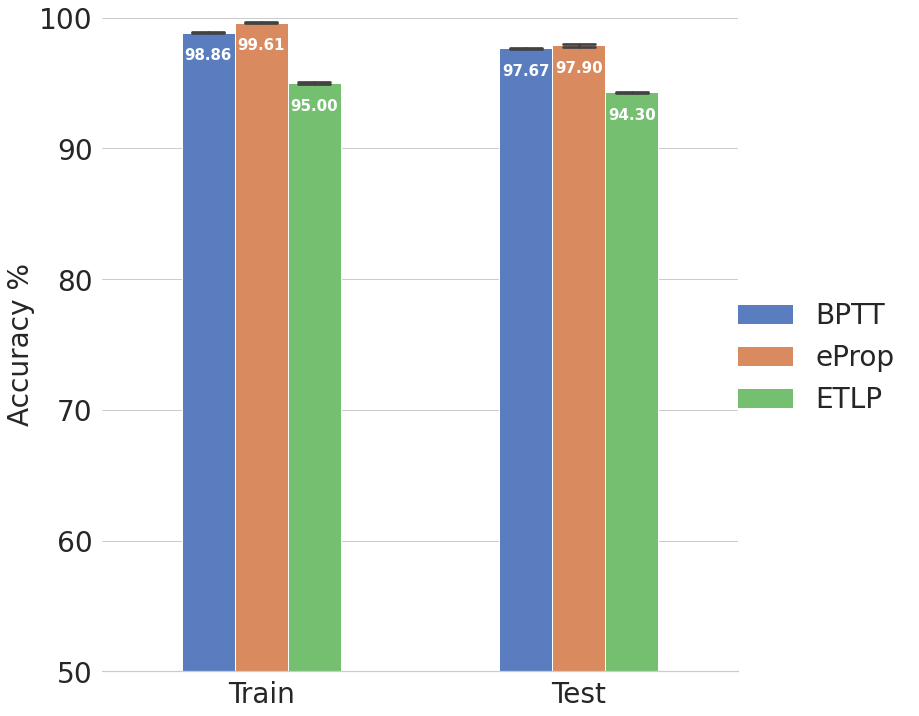} \label{fig:accuracy_nmnist_bar}}
\begin{minipage}[b]{0.5\textwidth}
\centering
\subfloat[]{\includegraphics[width=0.6\textwidth]{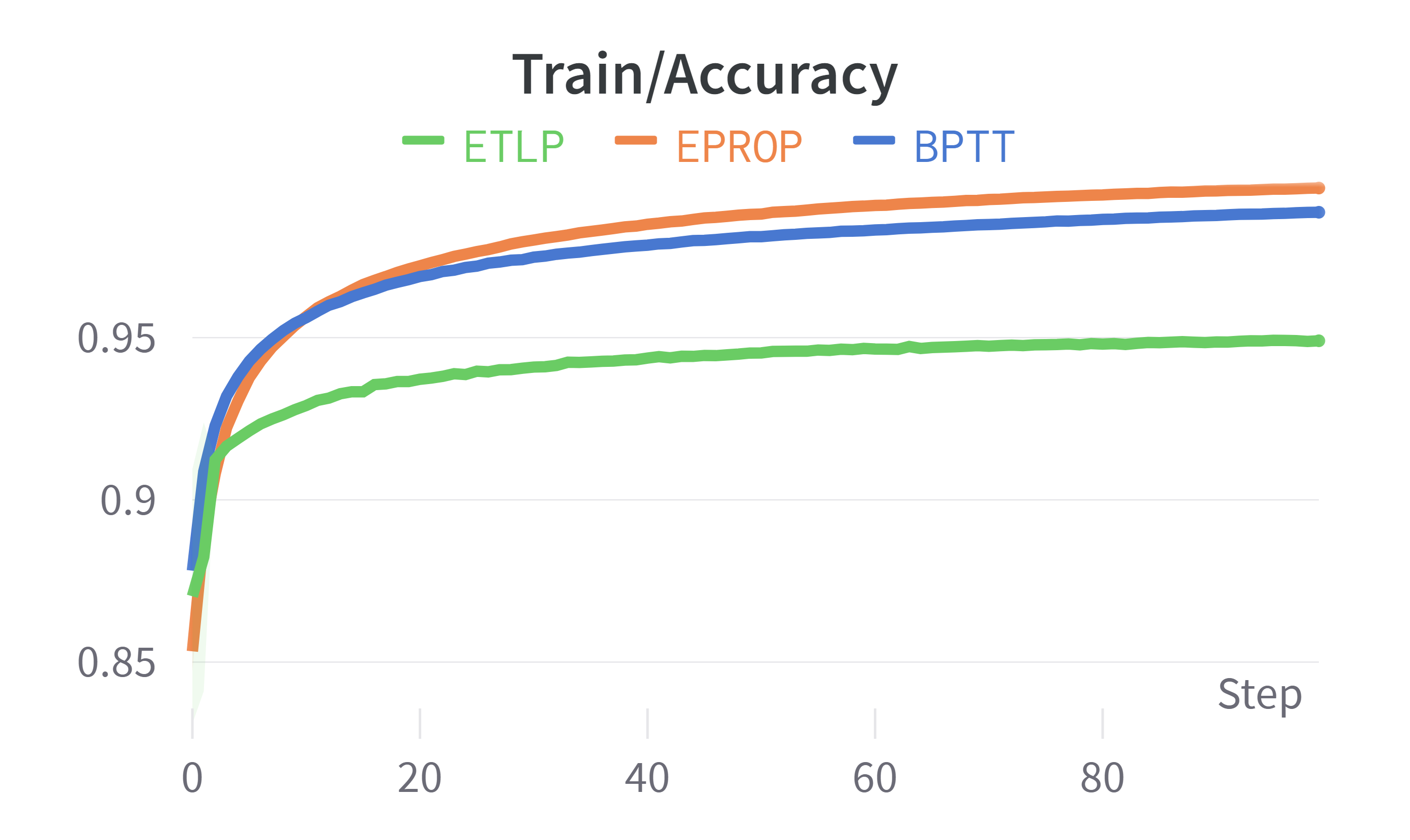} \label{fig:accuracy_nmnist_train}} \\
\subfloat[]{\includegraphics[width=0.6\textwidth]{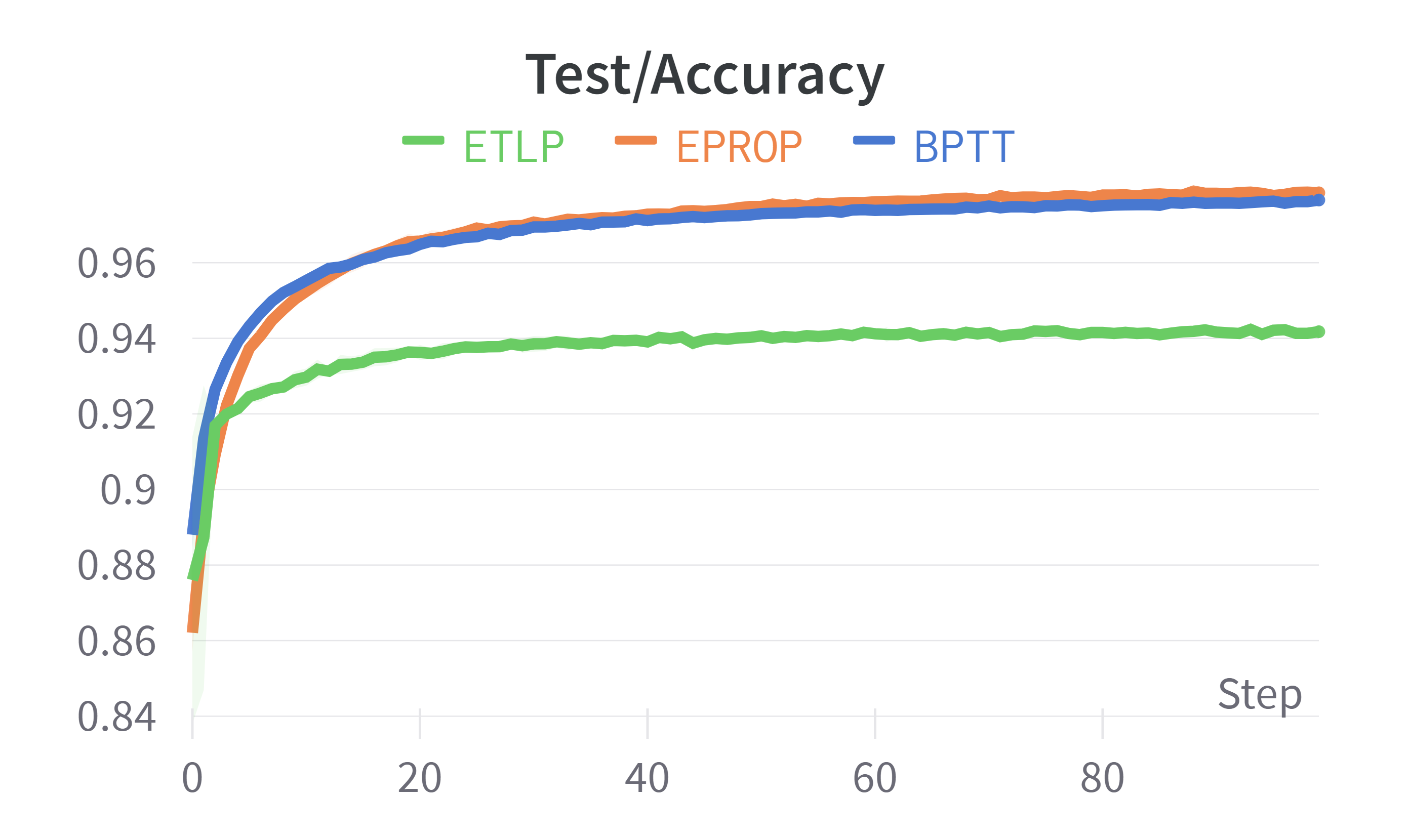} \label{fig:accuracy_nmnist_test}}
\end{minipage}
\caption{Accuracy of BPTT, eProp and ETLP on \ac{N-MNIST} using a feedforward network of \ac{LIF} neurons. Fig. (a) shows a comparison in accuracy between the training and test datasets, Figs. (b) and (c) show the accuracy over all the epochs for training and test, respectively.}
\label{fig:accuracy_nmnist}
\end{figure*}

The results on \ac{N-MNIST} dataset are shown in Figs. \ref{fig:accuracy_nmnist} and \ref{fig:accuracy_nmnist_bar}, for training and test set respectively. \ac{ETLP} shows an accuracy of 94.30\% on the test set, compared to 97.90\% and 97.67\% for eProp and \ac{BPTT} respectively.
\ac{BPTT} results are on par with the state-of-the-art with a similar topology \cite{Perez-Nieves_etal21,Bouanane_etal22}, while eProp reaches a lightly higher accuracy which is probably due to hyper-parameters.
On the other hand, \ac{ETLP} loses 3.60\% compared to eProp.

% ----------------- SHD ----------------- %

\subsection{SHD}
For the \ac{SHD} dataset, since it has more temporal dependency than \ac{N-MNIST} \cite{Perez-Nieves_etal21}, we tested both feedforward and recurrent networks with all-to-all explicit recurrent connections in the hidden layer. 
Both networks used 700 input neurons, 450 hidden neurons and 20 output neurons. Two different neuron models were explored: \ac{LIF} and \ac{ALIF}.

% ----------------- SHD fw ----------------- %

\begin{figure*}[!h]
\subfloat[]{\includegraphics[width=0.49\textwidth]{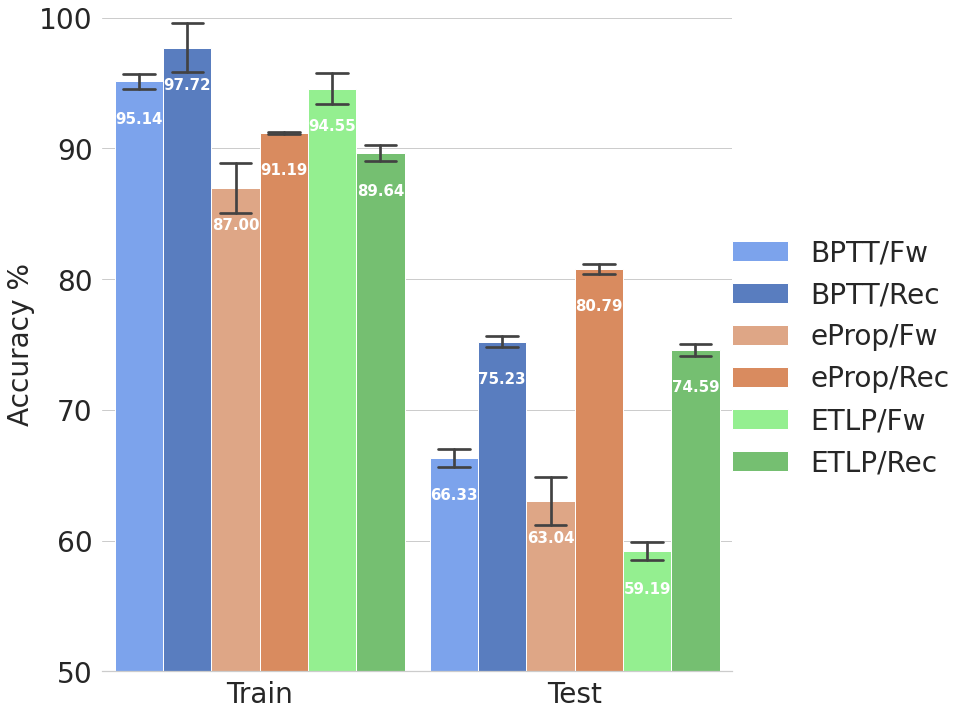} \label{fig:accuracy_shd_bar}}
\begin{minipage}[b]{0.5\textwidth}
\centering
\subfloat[]{\includegraphics[width=0.6\textwidth]{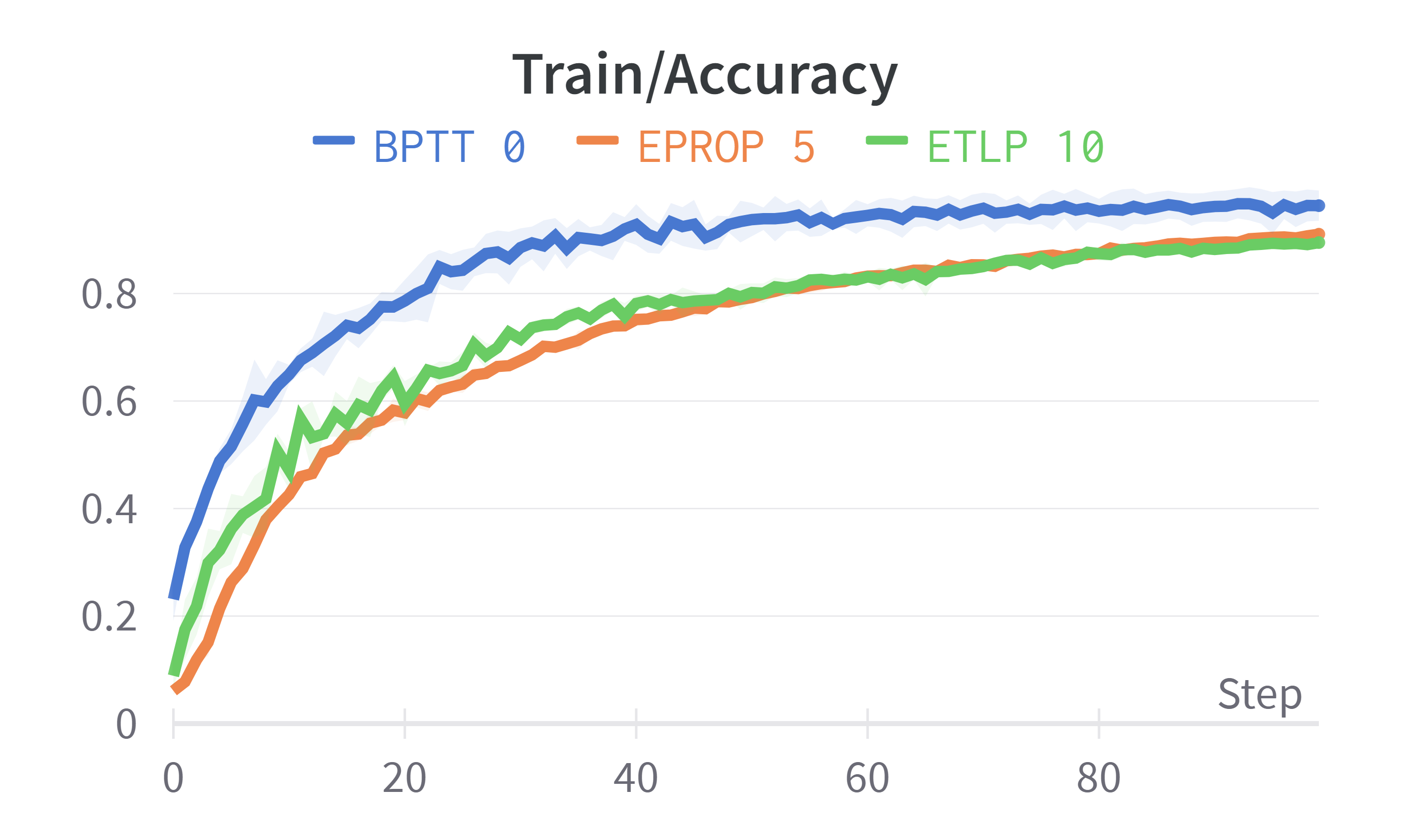} \label{fig:accuracy_shd_rec_train}} \\
\subfloat[]{\includegraphics[width=0.6\textwidth]{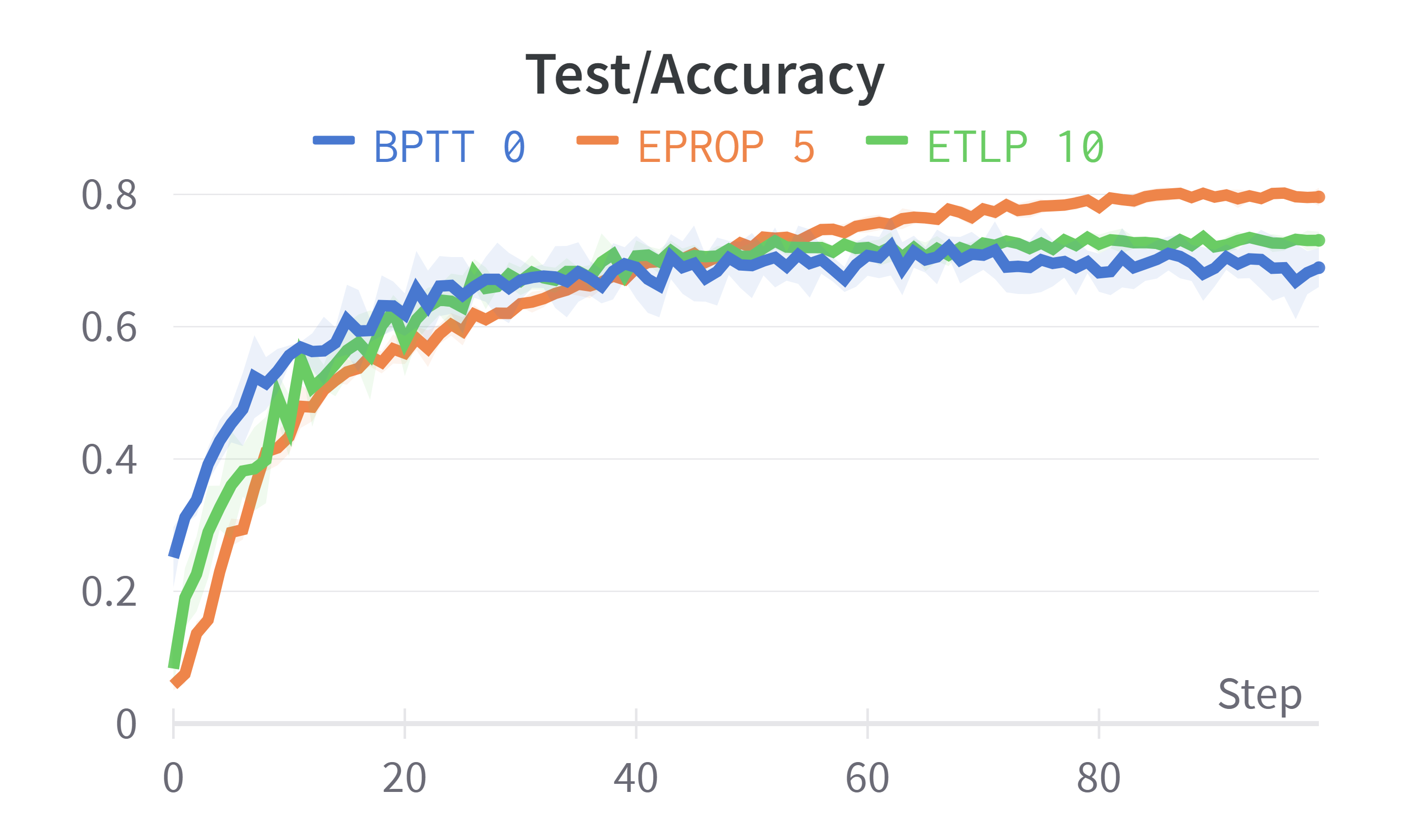} \label{fig:accuracy_shd_rec_test}}
\end{minipage}
\caption{Accuracy of BPTT, eProp and ETLP on \ac{SHD}. Fig. (a) shows a comparison in accuracy between the feedforward network with LIF neurons and the recurrent network with ALIF neurons. Figs. (b) and (c) show the accuracy of the the recurrent network for each learning rule on the training and test datasets, respectively, with the best $\theta$ value obtained for each one.}
\label{fig:accuracy_shd}
\end{figure*}

For the feedforward topology, a \ac{LIF} model was used. \ac{ETLP} has an accuracy of 59.19\%, compared to 66.33\% and 63.04\% for \ac{BPTT} and eProp respectively, as shown in Fig. \ref{fig:accuracy_shd}.

% ----------------- SHD rec ----------------- %

For the recurrent topology, an initial exploration over the $\theta$ parameter of the adaptive threshold has been made.
Several simulations have been performed for each rule and for each $\theta$ parameter (0, 5 and 10) to see how it affects the accuracy. Tab. \ref{tab:theta} shows the accuracy for each learning rule and $\theta$ parameter, where the $\theta$ value used for the final results is marked in green.

\begin{figure}[t]
    \centering
    \subfloat[]{\includegraphics[width=0.3\textwidth]{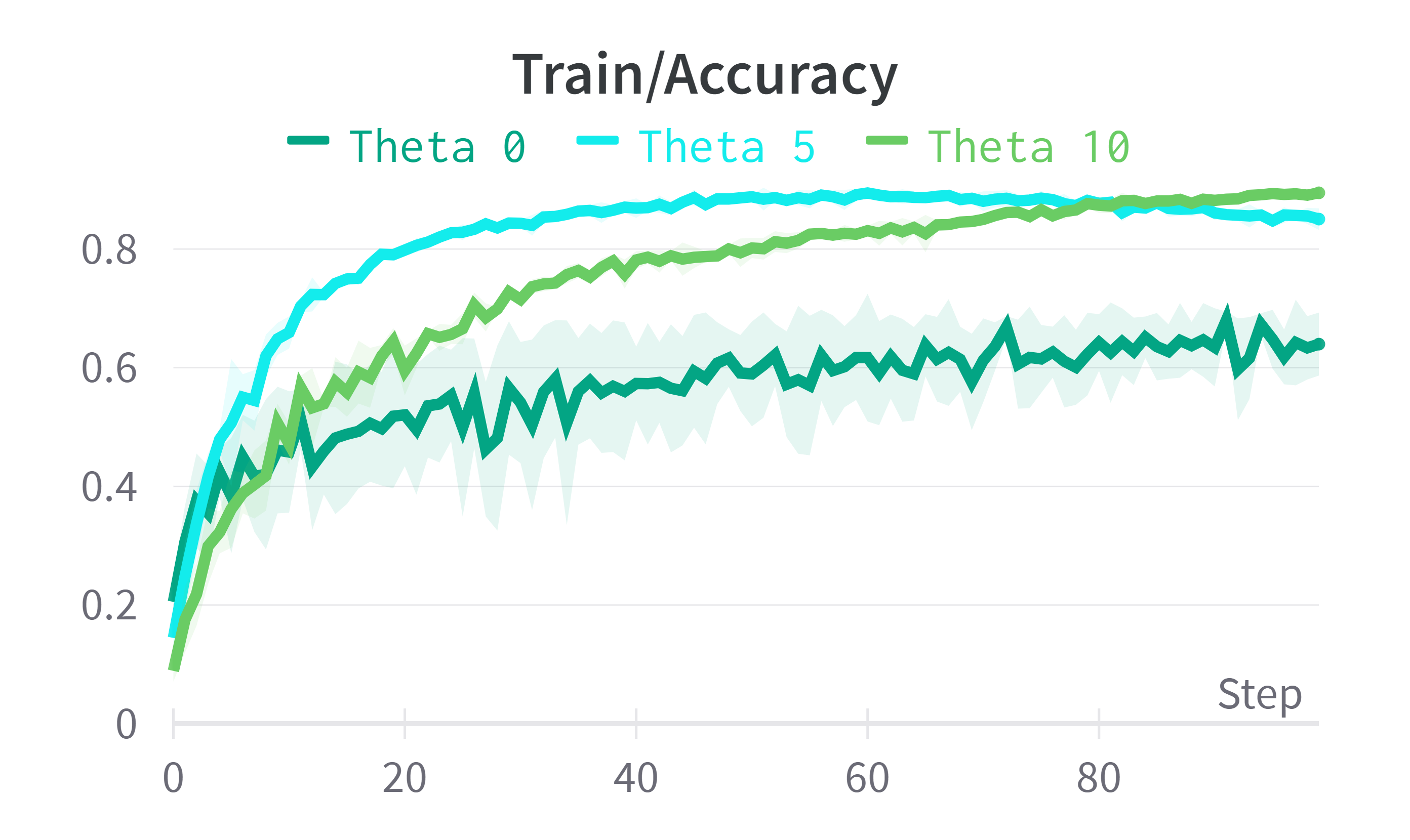} \label{fig:theta_etlpa}}
    \hfill
    \subfloat[]{\includegraphics[width=0.3\textwidth]{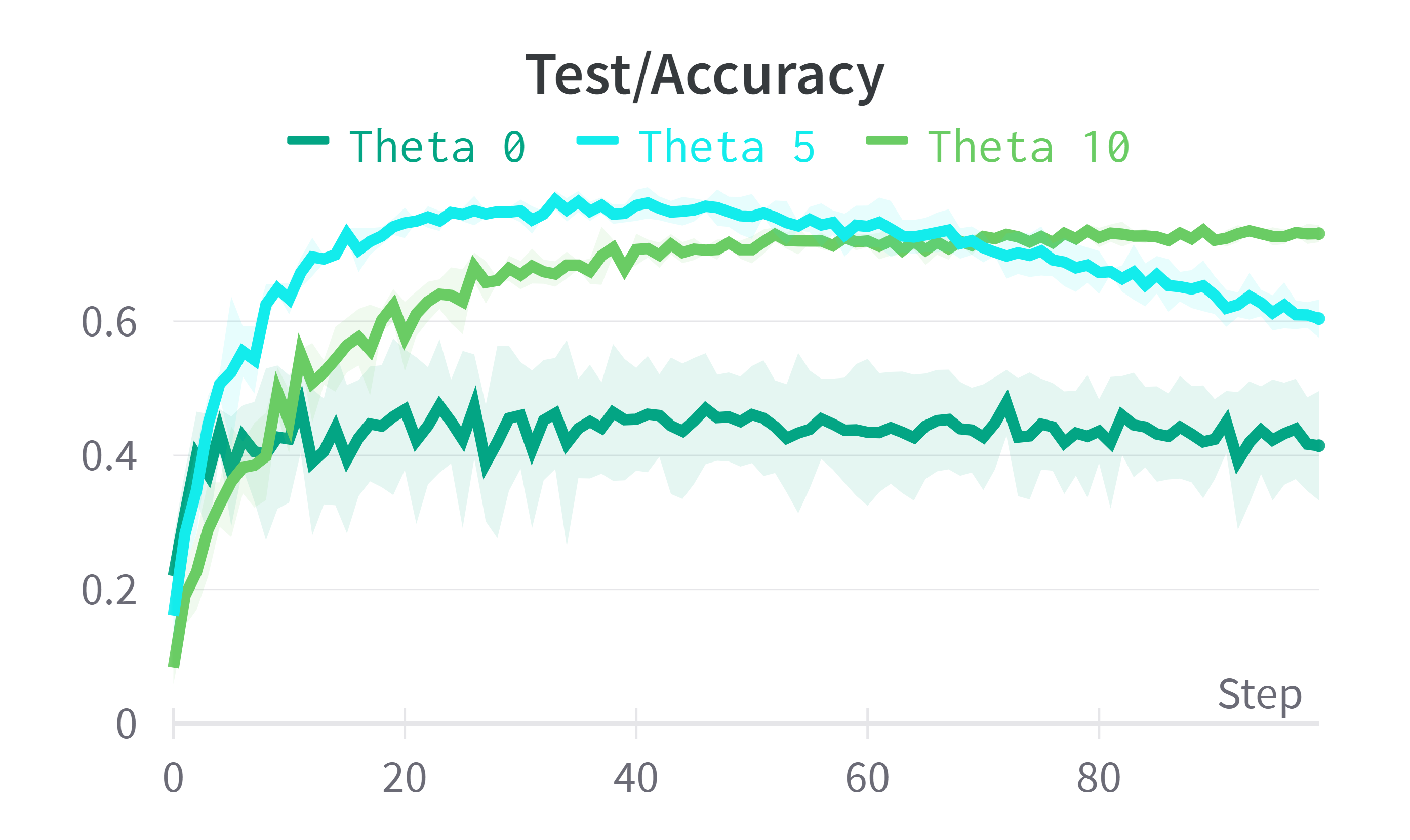} \label{fig:theta_etlpb}}
    \caption{Accuracy for different threshold adaptation on \ac{ETLP} in SHD dataset with the recurrent network.}
    \label{fig:theta_etlp}
\end{figure}

In the case of \ac{ETLP}, a higher maximum accuracy is obtained with $\theta=5$. However, it can be seen in Fig. \ref{fig:theta_etlp} that this accuracy then decreases with the epochs, going below the accuracy with $\theta=10$. Hence, $\theta=10$ is taken as the best value because it is more important to have a convergence with more data than to reach a maximum accuracy at a particular epoch.

\begin{table}
\caption{Test accuracy of $\theta$ parameter exploration on \ac{BPTT}, eProp and {ETLP}. On SHD dataset with a \ac{RSNN}.}
\label{tab:theta}
\begin{tabular}{|c|c|c|c|c|c|c|}
\hline
\textbf{Learning} & \multicolumn{2}{c|}{$\theta=0$} & \multicolumn{2}{c|}{$\theta=5$} & \multicolumn{2}{c|}{$\theta=10$}\\
\cline{2-7}
  \textbf{Method} & $\mu$  & $\sigma$ & $\mu$  & $\sigma$ & $\mu$ & $\sigma$ \\
\hline
BPTT  & \cellcolor{green!25} 75.23 & \cellcolor{green!25} 0.44 & 64.43 & 0.68 & 23.96 & 0.67 \\
\hline
eProp & 51.68 & 1.10 & \cellcolor{green!25} 80.79 & \cellcolor{green!25} 0.39 & 49.94 & 1.98 \\
\hline
ETLP  & 48.53 & 0.64 & 78.71 & 1.49 & \cellcolor{green!25} 74.59 & \cellcolor{green!25} 0.44 \\
\hline 
\end{tabular}
\end{table}

% --------- Theta parameters exploration --------- %

The final results obtained on \ac{SHD} dataset using a recurrent topology are shown in Fig. \ref{fig:accuracy_shd} based on the best $\theta$ parameter obtained per learning rule.
\ac{ETLP} ($\theta=10$) achieves a test accuracy of 74.59\%, showing a slight loss compared to \ac{BPTT} ($\theta=0$) that reaches 75.23\% and a consequent loss of 6.20\% compared to eProp that reaches 80.79\%.
We observe that adding explicit recurrence in the hidden layer of the network increases accuracy by a considerable amount of 8.90\% for \ac{BPTT}, 17.75\% for eProp and 15.40\% for \ac{ETLP}.
We also observe that the the \ac{BPTT} offline learning reaches a better accuracy without threshold adaptation (i.e. $\theta=0$), while it is required for the online learning of eProp (although spatially non-local) and \ac{ETLP}.

It is to note that, similar to \cite{Perez-Nieves_etal21}, we did not reach the state-of-the-art accuracy with \ac{BPTT} that reaches 80.41\% with a similar topology \cite{Bouanane_etal22}. 
This can be explained by the different training strategies where the output layer in \cite{Bouanane_etal22} is not spiking, as well as the chosen hyper-parameters that were not tuned specifically for \ac{BPTT}. 
Hence, \ac{BPTT} reaches in principle a better accuracy than both eProp and \ac{ETLP}.

% ------------------------------------------------------

\section{Hardware implementation}
\begin{figure}[!t]
    \centering
    \includegraphics[width=3.5in]{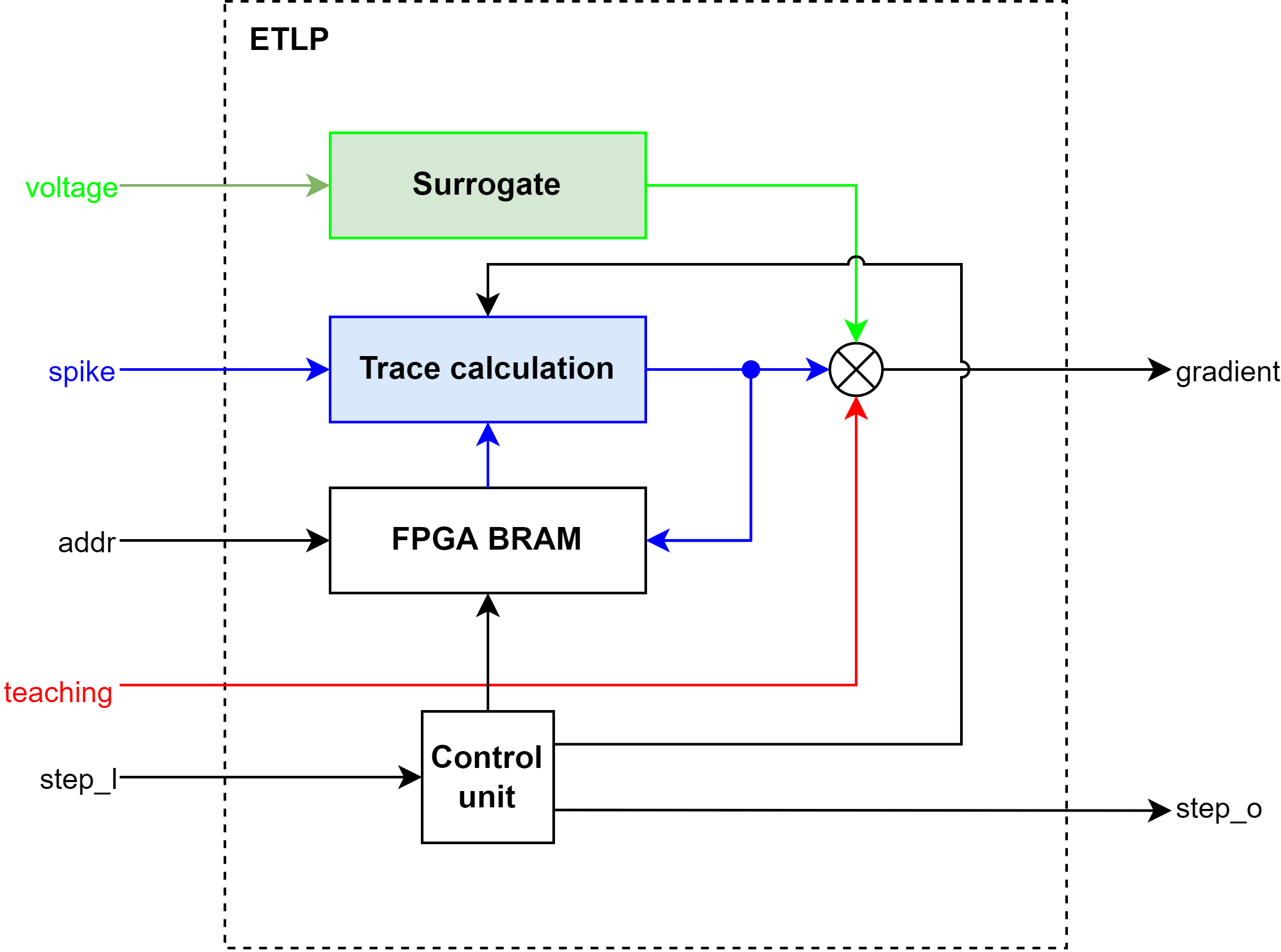}
    \caption{\ac{FPGA} implementation for \ac{ETLP} module. The control unit triggers the calculation of the new pre-synaptic spike trace value and stores it in memory. Once the gradient is calculated, it sends the control signal to the upper module so that it can read the calculated gradient value.}
    \label{fig:etlp_fpga}
\end{figure}

\noindent \ac{ETLP} could be easily implemented on a neuromorphic hardware, resulting in a low-latency and low-power consumption alternative for online learning. As a proof of concept, a \ac{FPGA} implementation of the \ac{ETLP} module for calculating the weight update based on \cite{Quintana2022} has been designed, as shown in Fig. \ref{fig:etlp_fpga}. The module receives the following signals as input for the gradient calculation: 

\begin{enumerate}
    \item The address of the pre-synaptic neuron that is connected to the synapse to be updated;
    \item The output of the pre-synaptic neuron (0 or 1) for the pre-synaptic spike trace calculation;
    \item The voltage value of the post-synaptic neuron for the surrogate voltage calculation;
    \item The teaching value, equal to the teaching neurons weights when they emit a spike which triggers the weight update.
\end{enumerate}

\begin{figure}[!t]
    \centering
    \includegraphics[width=3.5in]{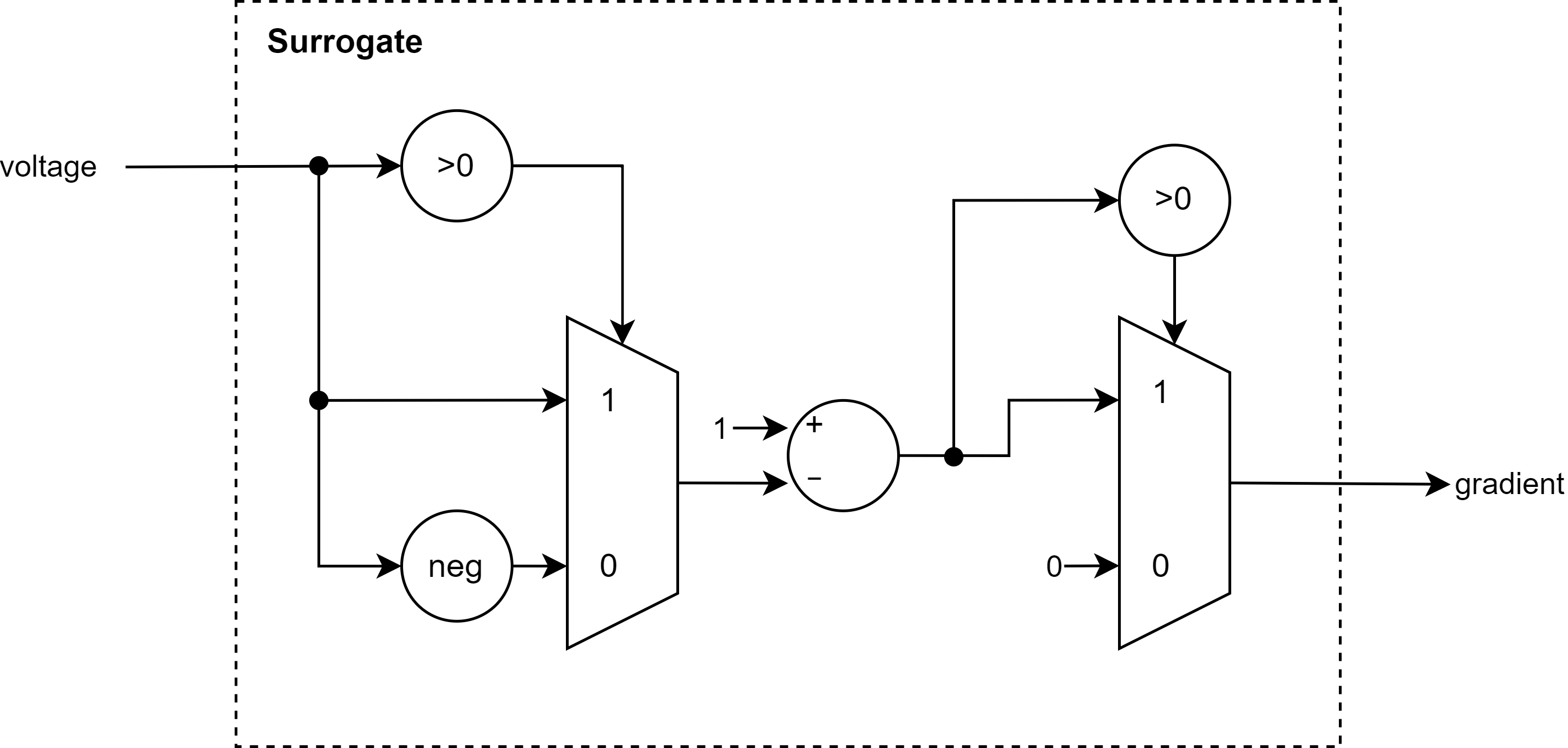}
    \caption{Triangular function surrogate gradient implementation on \ac{FPGA}.}
    \label{fig:fpga_surrogate}
\end{figure}

The implemented surrogate gradient shown in Fig. \ref{fig:fpga_surrogate}) is the triangular function, defined as $f(x) = max(0, 1 - |x|)$.

\begin{figure}[!t]
    \centering
    \includegraphics[width=3.5in]{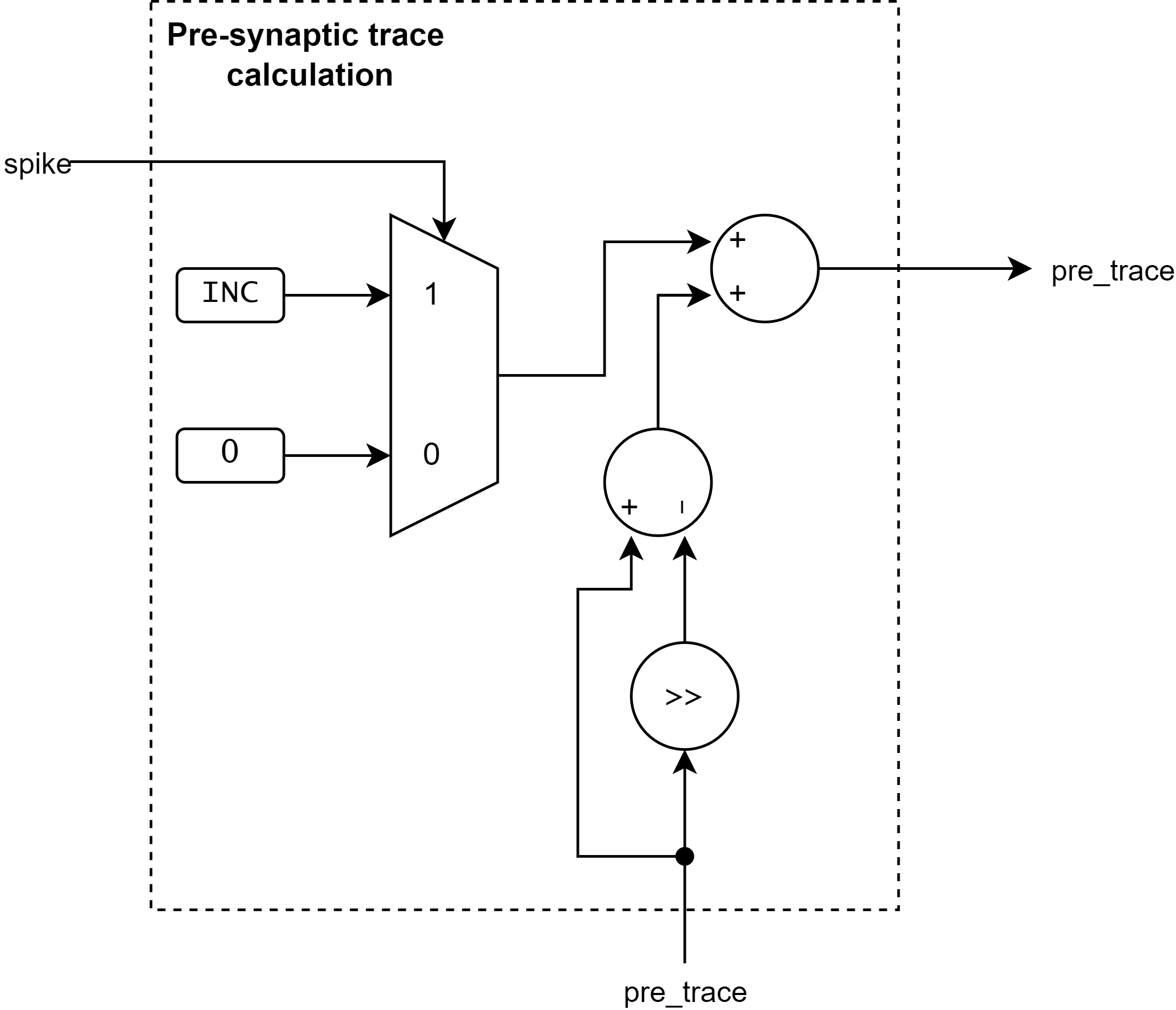}
    \caption{pre-synaptic spike trace calculation on \ac{FPGA}. In the case of a pre-synaptic spike arrives, a constant is added to the new pre-synaptic spike trace value.}
    \label{fig:fpga_trace}
\end{figure}

We assume to use some time-multiplexing handled by the upper module, but do not go into its architectural details as the \ac{ETLP} implementation is independent. 
The \ac{ETLP} module stores pre-synaptic spike trace values in a BRAM. Each time a step signal is received, the module retrieves the stored trace, calculates the new trace value and stores it back into memory. At the same time, it calculates the surrogate gradient from the value of the post-synaptic voltage. When both results are obtained, the final gradient is calculated with their product with the signal coming from the teaching neuron. Finally, the module produces a step signal to indicate to the upper module that it can update the corresponding synaptic weight.
This module can be optimized for asynchronous updates of the gradient (i.e. only when a spike from the teaching neurons is received).

\begin{figure}[!t]
    \centering
    \includegraphics[width=3.5in]{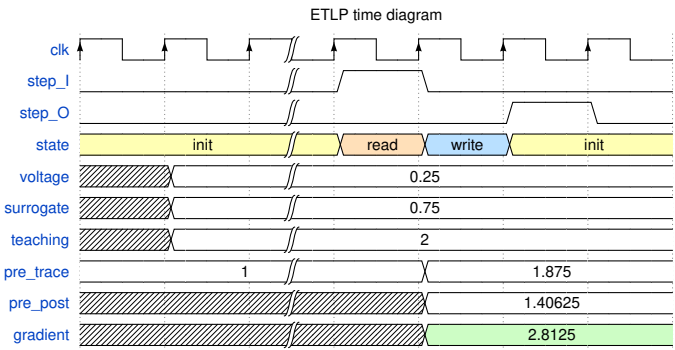}
    \caption{\ac{ETLP} \ac{FPGA} implementation time diagram.}
    \label{fig:time_diagram}
\end{figure}

This module takes three clock cycles to perform the calculation. Fig. \ref{fig:time_diagram} shows a time diagram of the process. The module would be in the ``initial" state until a signal from $step\_I$ is received. Then, it gathers the pre-synaptic spike trace from memory (``read" state), calculates the gradient value, and write the new pre-synaptic spike trace into memory (``write" state). Finally it produces an output signal to notify the upper module that the gradient value is available and changes the state back to ``init".

The module has been synthesised on a Nexys4 DDR (xc7a100tcsg324-1) \ac{FPGA} with the resource consumption shown in Tab. \ref{tab:fpga_resource}. A power estimation has been carried out at different clock frequencies. With a 100MHz clock, the device static and dynamic power consumption are approximately 97mW and 53mW respectively, where 47mW corresponds to I/O. With a 10MHz clock, the static power consumption is the same while the dynamic power consumption is reduced to 5mW, where 4.5mW corresponds to I/O. Finally, with a 1MHz clock, the dynamic power consumption is negligible, since the value is smaller than the resolution provided by the tool (below 1mW). 

Interestingly, when using one module per neuron, a 1MHz clock is sufficient to update a neuron in the hidden layer of our recurrent SNN with SHD, since the number of required clock cycles per second is (700 [synapses from input layer neurons] + 450 [synapses from hidden layer neurons) * 100 [updates per second] * 3 [clock cycles per updates] = 345,000, meaning a clock frequency of 345KHz. We can therefore expect a low-power consumption for the full network, especially when designing an \ac{ASIC} with a similar architecture.

\begin{table}
\begin{center}
\caption{Resource utilization of \ac{ETLP} module}
\label{tab:fpga_resource}
\begin{tabular}{|c|c|c|c|}
\hline
\textbf{Resource} & \textbf{Used} & \textbf{Available} & \textbf{Util\%} \\
\hline
Slice LUTs & 115 & 63400 & 0.18 \\
\hline
Flip Flop & 22 & 126800 & 0.02\% \\
\hline
DSPs & 2 & 240 & 0.83\% \\
\hline
BRAM tile & 1 & 270 & 0.37\% \\
\hline
\end{tabular}
\end{center}
\end{table}

% ------------------------------------------------------

\section{Discussion and conclusion}
\noindent In this work, we have proposed the \ac{ETLP} rule, a local learning rule in time and space which is triggered asynchronously based on the availability of the target or label. These features satisfy the hardware constraints of neuromorphic chips where computing and memory are co-localized.
We compared \ac{ETLP} to eProp (non-local in space) and \ac{BPTT} (non-local in time and space), and achieved a test accuracy of 94.30\% on \ac{N-MNIST} with a feedforward network of \ac{LIF} neurons and 74.59\% on \ac{SHD} with a recurrent network of \ac{ALIF} neurons. 
\ac{ETLP} has a loss of accuracy of 3.60\% and 6.20\% compared to the best accuracy of eProp on \ac{N-MNIST} and SHD, respectively, which is reasonable due to its fully event-driven and local plasticity paradigm.

Two features have a major impact on the accuracy when learning \ac{SHD} spatio-temporal patterns that have a rich temporal structure.
First, adding an all-to-all explicit recurrence in the hidden layer reduces training over-fitting and highly increases test accuracy by 8.90\%, 17.75\% and 15.40\% for \ac{BPTT}, eProp and \ac{ETLP}, respectively. This confirms the results reported on \ac{BPTT} in \cite{Bouanane_etal22}.
Second, the threshold adaptation of the \ac{ALIF} improves the accuracy of \ac{ETLP} and eProp by 26.06\% and 29.11\%, respectively, while it decreases the accuracy of \ac{BPTT} by 10.80\%. Although we did not extensively explore the reasons behind this behavior, it is clear for \ac{ETLP} that the threshold adaptation provides an online dynamic mechanism (in addition to the refractory period) to limit the neurons firing rate. In \ac{BPTT}, it might not be needed since the error signal is explicitly back-propagated and used in the weight updates.
An additional mechanism to explore for \ac{ETLP} to further reduce the firing rate and potentially gain energy-efficiency is the voltage regularization based on additional plasticity mechanisms such as the homeostatic membrane potential dependent synaptic plasticity \cite{Albers_etal16}.

\begin{table}
\begin{center}
\caption{Complexity analysis of the gradient computation over different learning rules. N represents the number of input neurons; M the number of neurons in a layer; T the length of the back-propagated sequence; Nr the number of readout neurons in DECOLLE; p the ratio of connected neurons/total possible.}

\label{tab:complexity}
\begin{tabular}{|c|c|c|}
\hline
\textbf{Learning} & \textbf{Space} & \textbf{Time} \\
\hline
BPTT & $O(NT)$ & $O(pNMT)$ \\
\hline
e-Prop & $O(pNM)$ & $O(pNM)$ \\
\hline
DECOLLE \cite{Kraiser_etal20} & $O(1)$ & $O(MNr + pNM)$ \\
\hline
ETLP & $O(1)$ & $O(pNM)$ \\
\hline
\end{tabular}
\end{center}
\end{table}

In addition to the accuracy benchmarking, we demonstrated a proof of concept hardware implementation on FPGA to show how it is possible to map the local computational primitives of \ac{ETLP} on digital neuromorphic chips.
It is also a way to better understand the simplicity of the plasticity mechanism in hardware, where the required information are two Hebbian pre-synaptic (spike trace) and post-synaptic (voltage put in a simple function) factors \cite{Khacef_etal22} and a third factor in the form of an external signal provided by the target or label.
Therefore, \ac{ETLP} is compatible with neuromorphic chips such as Loihi2 which supports on-chip three-factor local plasticity \cite{Davies_etal18,Orchard_etal21}.
Table \ref{tab:complexity} shows a comparison of the computational complexity of \ac{BPTT}, eProp, DECOLLE \cite{Kraiser_etal20} and \ac{ETLP}. 
We can see that \ac{ETLP} is indeed local in space like DECOLLE (\ac{ETLP} uses \ac{DRTP} while DECOLLE uses local losses) and local in time like eProp (although \ac{ETLP} needs one spike trace per pre-synaptic neuron only, while eProp needs in addition an eligibility trace per synapse), thus being more computationally efficient than previously proposed plasticity mechanisms.

\ac{ETLP} builds on previous developments toward more local learning mechanisms, and proposes a fully-local synaptic plasticity model that fits with the hardware constraints of neuromorphic chips for online learning with low-energy consumption and real-time interaction.
We have shown a good classification accuracy on visual and auditory benchmarks, both starting from a randomly initialized network to show a proof of convergence.
Future works will focus on implementing \ac{ETLP} on a more realistic experimental setup with few-shot learning. In order to adapt to new classes that have not been seen before, new labels will be provided to the network to automatically trigger plasticity. The open question is how to design the output layer, either by having more neurons that needed at the beginning or by adding new neurons when needed.
Since few-shot learning is based on very few labeled samples, we can train both initial network synaptic weights and \ac{ETLP} hyper-parameters based on the \ac{MAML} that has been recently applied to \acp{SNN} \cite{Stewart_Neftci_22}.

% ------------------------------------------------------

\section*{Acknowledgements}
\label{acknowledgements}
Fernando M. Quintana would like to acknowledge the Spanish \textit{Ministerio de Ciencia, Innovación y Universidades} for the support through FPU grant (FPU18/04321).
This work was partially supported by the Spanish grants (with support from the European Regional Development Fund) NEMOVISION (PID2019-109465RB-I00) and MIND-ROB (PID2019-105556GB-C33), the CogniGron research center and the Ubbo Emmius Funds of the University of Groningen.

% ------------------------------------------------------

\bibliographystyle{IEEEtranS}
\bibliography{main}

% ------------------------------------------------------

\clearpage

\begin{IEEEbiography}[{\includegraphics[width=1in,height=1in,clip,keepaspectratio]{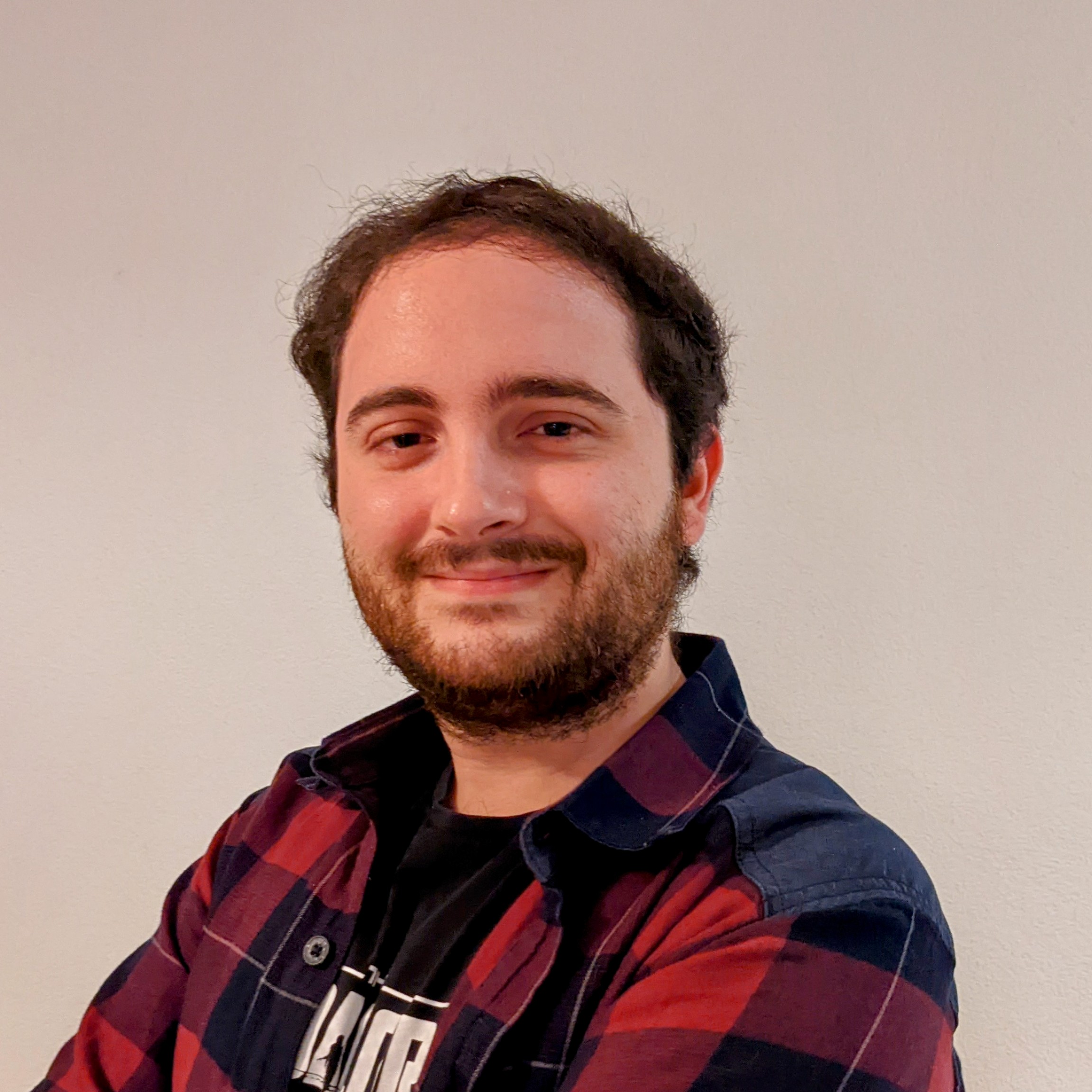}}]{Fernando M. Quintana} received the B.S. degree in Computer Engineering from the University of Cadiz (Spain), where he also obtained the master’s degree in Computer and Systems Engineering Research. He is currently pursuing his Ph.D. degree on local and online learning in Spiking Neural Networks.
\end{IEEEbiography}

\begin{IEEEbiography}[{\includegraphics[width=1in,height=1in,clip,keepaspectratio]{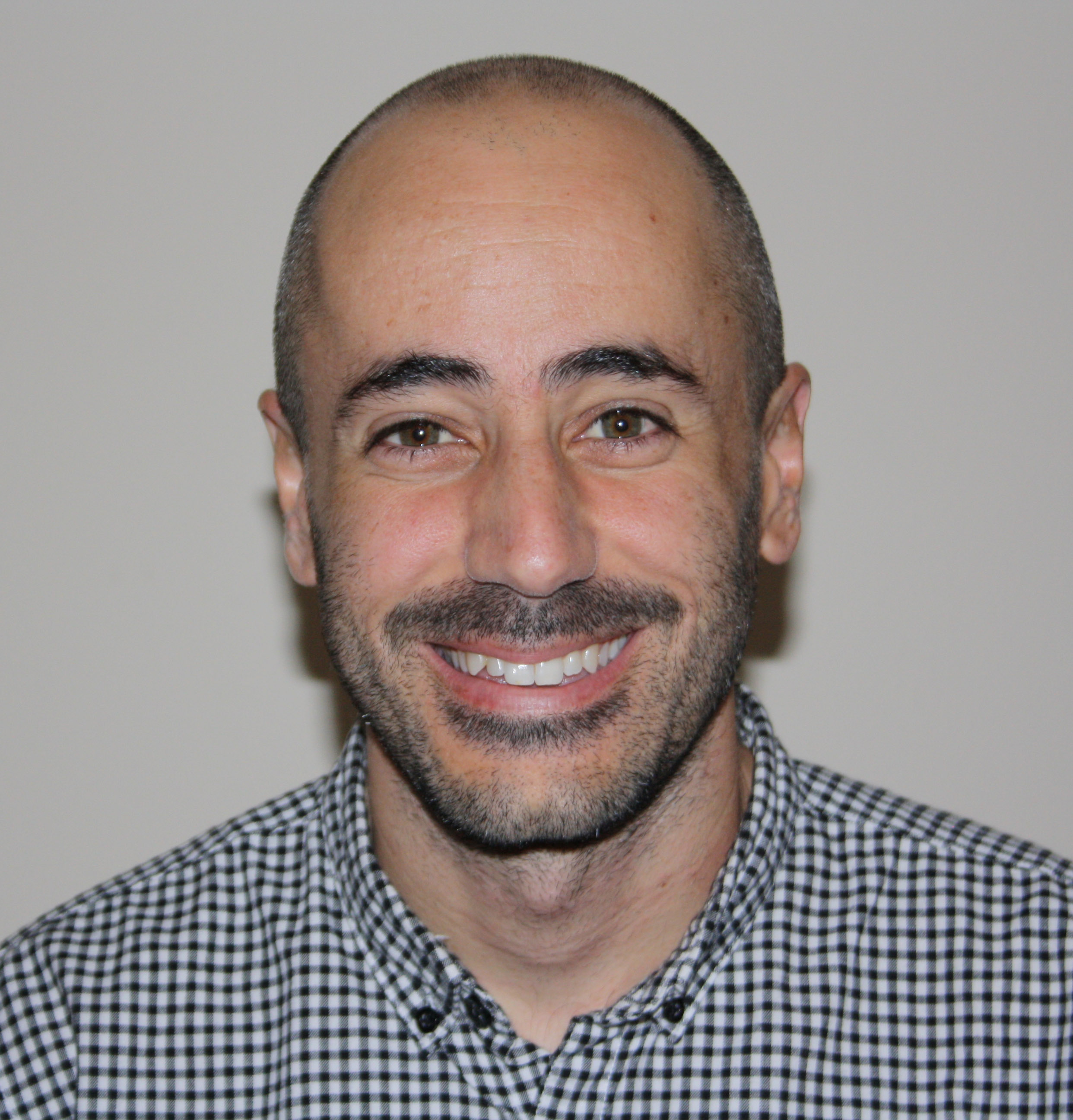}}]{Fernando Perez-Peña} received the Engineering degree in Telecommunications from the University of Seville (Spain) and his Ph.D. degree (specialized in neuromorphic motor control) from the University of Cadiz (Cadiz, Spain) in 2009 and 2014 respectively. In 2015 he was a postdoc at CITEC (Bielefeld University, Germany). He has been an Assistant Professor in the Architecture and Technology of Computers Department of the University of Cadiz since 2014. His research interests include neuromorphic engineering, CPG, motor control and neurorobotics.
\end{IEEEbiography}

\begin{IEEEbiography}[{\includegraphics[width=1in,height=1in,clip,keepaspectratio]{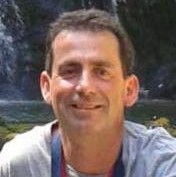}}]{Pedro L. Galindo} is Professor of Computer Science and Engineering and leader of the Intelligent Systems group at the University of Cadiz, Spain. His main research interest is the the application of Artificial Intelligence to the solution of real-world problems in industrial environments, especially those arised in Industry 4.0.
\end{IEEEbiography}

\begin{IEEEbiography}[{\includegraphics[width=1in,height=1in,clip,keepaspectratio]{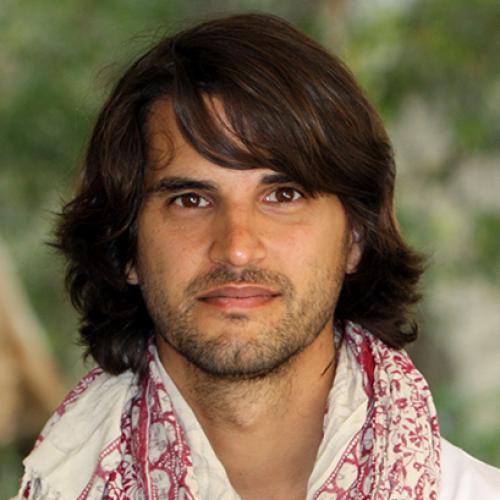}}]{Emre O. Neftci} received his MSc degree in physics from EPFL in Switzerland, and his Ph.D. in 2010 at the Institute of Neuroinformatics at the University of Zurich and ETH Zurich. He is currently an institute director at the Jülich Research Centre and Professor at RWTH Aachen. His current research explores the bridges between neuroscience and machine learning, with a focus on the theoretical and computational modeling of learning algorithms that are best suited to neuromorphic hardware and non-von Neumann computing architectures. 
\end{IEEEbiography}

\begin{IEEEbiography}[{\includegraphics[width=1in,height=1in,clip,keepaspectratio]{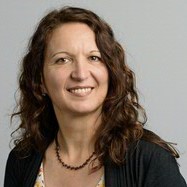}}]{Elisabetta Chicca} received a ``Laurea'' degree (M.Sc.) in Physics from the University of Rome 1 ``La Sapienza'', Italy in 1999 with a thesis on CMOS spike-based learning. In 2006 she received a Ph.D. in Natural Science from the Swiss Federal Institute of Technology Zurich (ETHZ, Physics department) and in Neuroscience from the Neuroscience Center Zurich. E. Chicca has carried out her research as a Postdoctoral fellow (2006-2010) and as a Group Leader (2010-2011) at the Institute of Neuroinformatics (University of Zurich and ETH Zurich) working on development of neuromorphic signal processing and sensory systems. From 2011 to 2020 she lead the Neuromorphic Behaving Systems (NBS) research group at Bielefeld University (Faculty of Technology and Cognitive Interaction Technology Center of Excellence, CITEC). She is currently the Chair of the Bioinspired Circuits and Systems (BICS) Laboratory, Faculty of Science and Engineering, University of Groningen, Groningen, The Netherlands. She has a long-standing experience with developing event-based neuromorphic systems and their application to biologically inspired computational models. Her research activities cover models of cortical circuits for brain-inspired computation, learning in spiking neural networks, bio-inspired sensing, and motor control.
\end{IEEEbiography}

\begin{IEEEbiography}[{\includegraphics[width=1in,height=1in,clip,keepaspectratio]{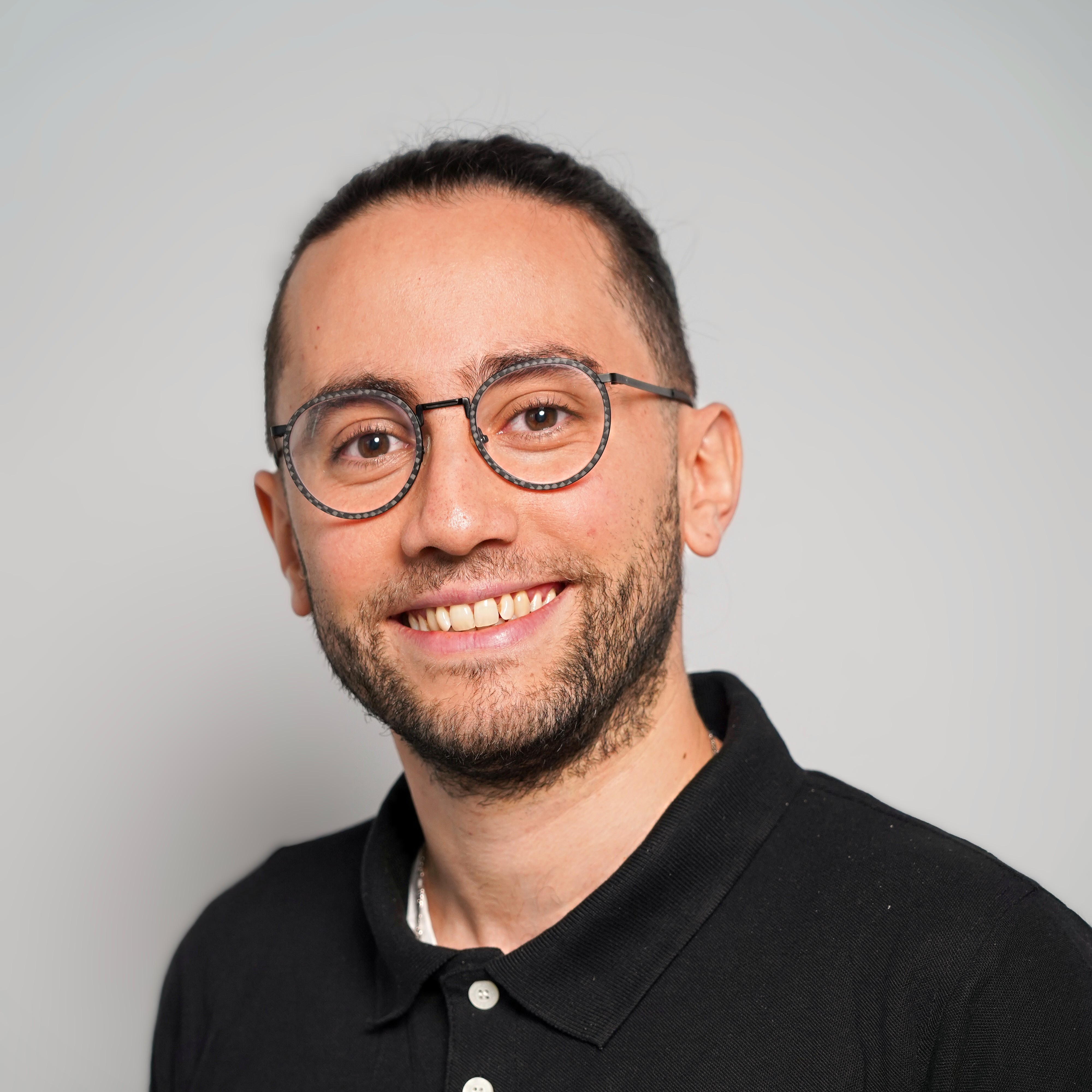}}]{Lyes Khacef} received his MSc in embedded systems from the University of Nice Sophia Antipolis (France). He pursued a PhD at Université Cote d’Azur (France) in brain-inspired computing where he worked on self-organizing neuromorphic architectures on FPGA for multimodal unsupervised learning based on local structural and synaptic plasticity. He then started a Postdoc at the University of Groningen (the Netherlands) where he is working on the modeling and hardware prototyping of spike-based local synaptic plasticity mechanisms for online learning of spatio-temporal patterns at the edge.
\end{IEEEbiography}

% ------------------------------------------------------

\vfill

\end{document}